\documentclass{article}

\usepackage{microtype}
\usepackage{graphicx}
\usepackage{subfigure}
\usepackage{booktabs} % for professional tables
\usepackage{wrapfig}

\usepackage{hyperref}

\usepackage{color,soul}

\usepackage{enumitem}% http://ctan.org/pkg/enumitem
\usepackage{float}

\usepackage{amsmath,amsfonts,bm}

\newcommand{\newterm}[1]{{\bf #1}}

\def\figref#1{figure~\ref{#1}}
\def\eqref#1{equation~\ref{#1}}
\def\1{\bm{1}}

\DeclareMathAlphabet{\mathsfit}{\encodingdefault}{\sfdefault}{m}{sl}
\SetMathAlphabet{\mathsfit}{bold}{\encodingdefault}{\sfdefault}{bx}{n}

\def\gC{{\mathcal{C}}}

\def\gM{{\mathcal{M}}}
\def\gN{{\mathcal{N}}}

\def\gU{{\mathcal{U}}}

\def\gX{{\mathcal{X}}}
\def\gY{{\mathcal{Y}}}
\def\gZ{{\mathcal{Z}}}

\newcommand{\E}{\mathbb{E}}

\newcommand{\R}{\mathbb{R}}

\newcommand{\KL}{D_{\mathrm{KL}}}

\usepackage[accepted]{icml2023}

\usepackage{amsmath}
\usepackage{amssymb}
\usepackage{mathtools}
\usepackage{amsthm}

\usepackage[capitalize,noabbrev]{cleveref}

\theoremstyle{plain}
\newtheorem{theorem}{Theorem}[section]
\newtheorem{proposition}[theorem]{Proposition}

\theoremstyle{definition}

\theoremstyle{remark}

\usepackage[textsize=tiny]{todonotes}

\icmltitlerunning{Submission and Formatting Instructions for ICML 2023}

\begin{document}

\twocolumn[
\icmltitle{Coupled Variational Autoencoder}



\icmlsetsymbol{equal}{*}

\begin{icmlauthorlist}
\icmlauthor{Xiaoran Hao}{Rutgers}
\icmlauthor{Patrick Shafto}{Rutgers,IAS}
\end{icmlauthorlist}

\icmlaffiliation{Rutgers}{Department of Mathematics and Computer Science, Rutgers University-Newark, New Jersey, USA}
\icmlaffiliation{IAS}{School of Mathematics, Institute for Advanced
Study, New Jersey, USA}

\icmlcorrespondingauthor{Xiaoran Hao}{xh197@rutgers.edu}

\icmlkeywords{Machine Learning, ICML}

\vskip 0.3in
]



\printAffiliationsAndNotice{} % otherwise use the standard text.

\begin{abstract}
Variational auto-encoders are powerful probabilistic models in generative tasks but suffer from generating low-quality samples which are caused by the holes in the prior. We propose the Coupled Variational Auto-Encoder (C-VAE), which formulates the VAE problem as one of Optimal Transport (OT) between the prior and data distributions. The C-VAE allows greater flexibility in priors and natural resolution of the prior hole problem by enforcing coupling between the prior and the data distribution and enables flexible optimization through the primal, dual, and semi-dual formulations of entropic OT. 
Simulations on synthetic and real data show that the C-VAE outperforms alternatives including VAE, WAE, and InfoVAE in fidelity to the data, quality of the latent representation, and in quality of generated samples. 
\end{abstract}

\section{Introduction}
The combination of variational Bayesian inference and deep latent variable models resulted in one of the most powerful generative models, Variational autoencoders (VAEs) \citep{Kingma13,Rezende14}. 
Scaleable inference and simple, flexible latent representations have enabled VAEs to make substantial progress in fields including image and video generation \citep{Razavi19,yan2021videogpt}, audio and music synthesis \citep{dhariwal2020jukebox,Kim2021ConditionalVA}, molecular processes \citep{Lim2018MolecularGM},  semi-supervised learning \citep{Kingma14, Izmailov}, and unsupervised representation learning \citep{Fortuin2018SOMVAEID, Oord2017NeuralDR}.

VAEs assume a latent prior over a generic space, and a variational approximate posterior, which in principle allow flexibility in modeling domains. 
However, for simplicity, an isotropic multivariate Gaussian is often employed for both prior distribution and variational posterior. A too simplistic prior could lead to over-regularization and a too simplistic family of variational posterior distributions leaves a big gap between true posterior and variational posterior, which limits performance on complex data \citep{Burda2016ImportanceWA,Kingma16,DaiW19}. 
Although people have proposed many methods to solve these problems. i.e enriching the variational family \citep{Rezende15,Kingma16}, using a more flexible \citep{tomczak18a,implicitprior,Casale2018GaussianPP} or non gaussian prior \citep{s-vae18,d-vae}, 
these approaches remain restricted to parametric distributions. 
A second limitation is the  ``prior hole problem'' which refers to the mismatch between prior and aggregate posterior that reduces the quality of ancestral samples \citep{Rezende18,Aneja2021}. The problem arises due to the variational approximation and density estimation, which when imperfect allows ``holes''---areas where the aggregate posterior has low density compared to the prior---where the decoder has not been trained but has a high probability of being generated under the prior. 

%^because the objective of VAEs, as Bayesian inference, does not impose any constraint on the aggregate posterior which should match the prior when achieving the optimum. \pat{??} If the inference is accurate and the model really learns the data distribution, there will be no gap between the prior and aggregate posterior. This undesired phenomenon will hurt the quality of samples as such holes usually have relatively low density under aggregate  posterior if compared to that of prior. This means the decoder or generator is rarely trained in these areas but gets sampled frequently if it has high density under prior.

Optimal transport (OT) \citep{Villani2008OptimalTO,kantorovich,cuturi13} provides a cogent solution to both the restriction to parametric posterior distributions and the mismatch between the prior and aggregate posterior distributions. 
Kantorovich OT computes optimal couplings between marginal distributions, and entropic OT allows efficient computation. 
%OT is originally from mathematics and economics  has been introduced into the machine learning field for many years \citep{Villani2008OptimalTO,kantorovich,cuturi13}. 
%\pat{FIXME: focus the rest on solving our problem.}
%\pat{This isn't quite relevant to our problem:} It defines a standard way to lift a metric on the ground space to the space probability measures. \pat{???} 
%OT defines a metric between two arbitrary probability measures given a pre-defined cost function. However, it suffers from its expensive computation burdens. It was shown that this cost can be largely mitigated by settling for a cheaper approximation obtained through entropy regularization \citep{cuturi13} Moreover, the EOT loss coincides with ELBO when using a specific cost function. It broadens our perspective on probabilistic modeling and gives us more intuition from transportation theory.
Computational methods for solving EOT problems do not require parametric distributions, which in principle allows greater flexibility. 
Moreover, coupling the marginals---here the prior and the empirical distribution of the data---can resolve the prior hole problem. 

We propose the Coupled VAE (C-VAE), which generalizes previous VAEs and addresses their limitations. 
%Building on InfoVAE, we formulate the optimization problem as entropic OT by considering the 
%a generalization of the VAE, from entropy-regularized optimal transport (EOT) perspective. Instead of optimizing VAE objectives over all data points, we consider it as an expectation over empirical data distribution and optimize the joint distribution over latent variables and data.  This allows us to reformulate it as an EOT problem.  
We derive an EOT-based algorithm for training the decoder and encoder. We use the dual and semi-dual OT formulation solving for the approximate posterior in the continuous prior case and the Sinkhorn algorithm in the discrete prior case. We can therefore work with an enriched family of approximate posterior distributions, and any distribution as prior without extra cost. The prior hole problem is resolved naturally via marginal constraints of EOT. We illustrate the flexibility in both prior and posterior, and the resolution of the prior hole problem through detailed simulations.  

 \newterm{Notation.} For a metric space $\gX$, $\gC(\gX)$ denotes the space of all continuous functions on $\gX$ and $\gM^1_+(\gX)$ denotes the set of positive Radon probability measures (i.e. of unit mass) on $\gX$. Upper cases $X$ denote random variables that take values on $\gX$. $X \sim \alpha$ says that a random variable $X$ follows a distribution $\alpha \in \gM^1_+(\gX$). Capital letters $P_X$ denote probability distribution and $p(x)$ to represent the probability density function. When there is no ambiguity, we use the same notation for distributions and their densities.

\section{Background}
We start by reviewing VAEs and the generalization to InfoVAE. Then we will introduce Entropy-regularized Optimal transport and the connection between VAEs and EOT.

\subsection{Variatonal autoencoders and InfoVAE}
Consider two compact metric spaces $\gX$ and $\gZ$. Latent variable generative models are usually defined in the form of a parametric model of joint distribution that admits density function $p_{\theta}(x,z) = p_{\theta}(x|z)p(z) $ over observable variable $x \in \gX$ and latent variable $z \in \gZ$. $p(z)$ is typically a simple prior distribution such as uniform or Gaussian. $p_{\theta}(x|z) $ is the conditional distribution parametrized by neural networks with parameters $\theta$. Marginalizing out the latent variable $z$ results in the model distribution $p_{\theta}(x) = \int p_{\theta}(x,z)dz =\int p_{\theta}(x|z)p(z)dz $. The goal of generative modeling is to maximize marginal likelihood:
\begin{equation} \label{gen_obj}
\E_{p_D(x)}[\log p_{\theta}(x)] = \E_{p_D(x)}[\log \E_{p(z)}[p_{\theta}(x|z)]],
\end{equation}
Where $p_D(x)$ represents data distribution.
However, because of the parameterization of $ p_{\theta}(x|z)$, the log marginal likelihood is intractable in general. VAEs maximize a surrogate, the evidence lower bound (ELBO) instead. We denote it by $\mathcal{L}_\mathbf{VAE}$:
\begin{equation} \label{vae_obj}
 \E_{p_D(x)}[\E_{q_{\phi}(z|x)}[\log p_{\theta}(x|z)] - D_{KL}(q_{\phi}(z|x)||p(z))], 
\end{equation}

where $q_{\phi}(z|x)$ is the variational posterior that is modeled by deep neural networks with parameter $\phi$. In the context of Auto-encoders, $p_{\theta}(x|z)$ and $q_{\phi}(z|x)$ are called decoder and encoder respectively. The first term of \eqref{vae_obj} is called the reconstruction term and the second term is called the regularization term. The typical choice of encoder distribution is Gaussian with a diagonal covariance matrix, i.e.,  $q_{\phi}(z|x) = \gN(z| \mu_\phi(x), \mathrm{diag}(\sigma_\phi^2(x)))$. The decoder distribution depends on the context, e.g. Bernoulli distribution if $x$ is binary and the Gaussian distribution if $x$ is continuous. 

The parameters of the neural networks are obtained by optimizing the ELBO. For continuous latent variables, this could be done efficiently through the re-parameterization trick \citep{Kingma13}. 

InfoVAE \citep{Zhao2019InfoVAEBL} generalizes the VAE family of models by introducing a scaling parameter to the KL divergence term and including a mutual information term that encourages high mutual information between $x$ and $z$. Formally, up to an additive constant, we can derive an equivalent expression of ELBO:
\begin{align} \label{vae_obj2}
 \mathcal{L}_\mathbf{VAE} = -&D_{KL}(q_{\phi}(z)||p(z))]  \nonumber \\ 
 &-\E_{q_{\phi}(z)}[D_{KL}(q_{\phi}(x|z)||p_{\theta}(x|z)] + \mathrm{const}.
\end{align}
where constant is given by $\E_{p_{D}(x)}[\log p_{D}(x)]$, $q_{\phi}(z) = \int q_{\phi}(z|x)p_D(x)dx $ and $q_{\phi}(x|z) =  \frac{q_{\phi}(z|x)p_{D}(x)}{q_{\phi}(z)} $. Then the loss of InfoVAE is derived as follows:
\begin{align} 
\mathcal{L}_\mathbf{InfoVAE} = &-\lambda D_{KL}(q_{\phi}(z)||p(z)) \nonumber \\ 
&- \mathbb{E}_{q_{\phi}(z)}[D_{KL}(q_{\phi}(x|z)||p_{\theta}(x|z))] \nonumber \\
&+ \alpha I_q(x;z) \nonumber \\ 
= &\mathbb{E}_{p_{D}(x)}\mathbb{E}_{q_{\phi}(z|x)}[\log p_{\theta}(x|z)] \nonumber \\ 
&- (1-\alpha)\mathbb{E}_{p_{D}(x)}[D_{KL}(q_{\phi}(z|x)||p(z))] \label{Eq:infovae}\\ 
&- (\alpha + \lambda -1)D_{KL}(q_{\phi}(z)||p(z)). \nonumber
\end{align}
When $\alpha = 0$ and $\lambda = 1$, InfoVAE recovers simple VAE. When $\alpha + \lambda - 1 = 0$, we have $\beta$-VAE \citep{Higgins2017betaVAELB}. If $\alpha=\lambda=1$ and KL divergence is replaced with Jensen Shannon divergence, and the model becomes adversarial autoencoder (AAE) \citep{Makhzani16}. 

The final optimization problem of InfoVAE is,
\begin{equation} \label{vae_obj3}
\min_{p_{\theta}(x|z)} \min_{q_{\phi}(z|x)} - \mathcal{L}_\mathbf{InfoVAE}. 
\end{equation}
Typically, optimizations over $p_{\theta}(x|z)$ and $q_{\phi}(z|x)$ are done jointly and over parameters $\theta$ and $\phi$. We introduce separated minimizations here in order to make connections in the following sections.

\newterm{Prior hole problem}. The problem refers to the situation when $q_{\phi}(z)\neq p(z)$ and there exist regions that have high density under $p(z)$ but low, possibly zero, density under $q_{\phi}(z)$. This does harm to the ancestral sampling process of the VAEs model as the samples drawn from the prior may not be decoded closely to the samples from the data distribution. In other words, the holes in the prior will generate low-quality samples. InfoVAE mitigates but does not solve, this problem to some degree as it has an explicit penalty term with a pre-defined weight parameter.  

\subsection{Entropy-regularized optimal transport}
For two continuous probability measures $\mu \in \gM_+^1(\gX)$ and $\nu \in \gM_+^1(\gY)$, given a measurable function $c(x,y): \gX \times \gY \rightarrow \R$ which represents the ground cost of moving a unit of mass from $x$ to $y$, 
Kantorovich OT \citep{kantorovich} seeks the optimal transport plan $\pi(x,y)$ subject to marginals $\mu$, $\nu$ with minimum transport loss. Formally, the  Kantorovich OT is, 
 \begin{align}
      OT(\mu,\nu) \stackrel{{\mathsf{def}}}{=} \min_{\pi \in \gU(\mu, \nu)} \int_{\gX\times \gY} c(x,y)d \pi(x,y).
 \end{align}
where the feasible set  $\ \gU(\mu,\nu)$ consists of all probability measures defined over the product space $\gX \times \gY$ with marginal measures $\mu$ and $\nu$ respectively.

Entropy regularized Optimal Transport (EOT) \citep{cuturi13, genevay16} includes an entropy regularization term into the original Kantorovich OT objective:

 \begin{align}
      OT_\varepsilon(\mu,\nu) \stackrel{{\mathsf{def}}}{=} \min_{\pi \in \gU(\mu, \nu)} \int_{\gX\times \gY} &c(x,y)d \pi(x,y) \nonumber \\ 
      &+ \varepsilon \KL(\pi \Vert \mu \otimes \nu). \label{Eq:EOT}
 \end{align}
where $\epsilon$ is the regularization weight and relative entropy $\KL(\pi \Vert \mu \otimes \nu)$ is defined as: 
\begin{align}
 \KL(\pi \Vert \mu \otimes \nu) \stackrel{{\mathsf{def}}}{=} \int_{\gX\times \gY} \log \left(\frac{d\pi(x,y)}{d\mu(x)d\nu(y)} \right) d\pi(x,y). 
\end{align}

By Fenchel-Rockafellar duality, the Kantorovich problem with entropy regularization admits dual formulation, which can be expressed as the maximization of an expectation:
\begin{align}
OT_\varepsilon(\mu,\nu)
&= \max_{\substack{u \in \gC(\gX) \\ v \in \gC(\gY)}} \int_{\gX}u(x)d\mu(x) + \int_{\gY}v(z)d\nu(y) \nonumber \\ 
&\qquad- \varepsilon \int_{\gX \times \gY}e^{\frac{u(x)+v(y)-c(x,y)}{\varepsilon}}d\mu(x)d\nu(y) \\
&= \max_{u \in \gC(\gX)} \int_{\gX}u(x)d\mu(x) + \int_{\gY}u^{c,\varepsilon}(y)d\nu(y) \nonumber \\
&= \max_{u \in \gC(\gX)} \E_{Y \sim \nu} [\int_{\gX}u(x)d\mu(x) + u^{c,\varepsilon}(Y) ],  \label{Eq:dual}
\end{align}

% \begin{align}
% OT_\varepsilon(\mu,\nu)
% &= \max_{u \in \gC(\gX), v \in \gC(\gY)} \int_{\gX}u(x)d\mu(x) \nonumber \\ 
% &\qquad\qquad + \int_{\gY}v(z)d\nu(y) \nonumber \\ 
% &\qquad\qquad - \varepsilon \int_{\gX \times \gY}e^{\frac{u(x)+v(y)-c(x,y)}{\varepsilon}}d\mu(x)d\nu(y) \\
% &= \max_{u \in \gC(\gX)} \int_{\gX}u(x)d\mu(x) + \int_{\gZ}u^{c,\varepsilon}(y)d\nu(y) \nonumber \\
% &= \max_{u \in \gC(\gX)} \E_{Y \sim \nu} \left[\int_{\gX}u(x)d\mu(x) + u^{c,\varepsilon}(Y) \right],  \label{Eq:dual}
% \end{align}

where $u^{c,\varepsilon}(y)$ is the $ c, \varepsilon$ -transform:
\begin{equation}
u^{c,\varepsilon}(y) \stackrel{{\mathsf{def}}}{=} -\varepsilon \log \int_{\gX}e^{\frac{u(x)-c(x,y)}{\varepsilon}}d\mu(x). 
\end{equation}

Sinkhorn's algorithm solves discrete EOT with linear convergence \citep{cuturi13}. However, EOT does not scale well to measures supported on a large number of points and it also assumes discrete measures. A fundamental property of the dual problem (Equation \ref{Eq:dual}) is stochastic gradient methods are applicable as long as we can sample from the marginal distributions \citep{genevay16,Seguy}. Dual variables can be parameterized by neural networks in continuous settings or left as a finite vector in discrete cases. This provides a method to apply optimal transport to large-scale machine learning tasks such as generative modeling.

The relationship between primal and dual problems allows switching between formulations. After solving for the dual variables, we can recover a feasible solution to the primal problem by the first-order optimality condition, 
\begin{equation}
d\pi(x,y) = \exp \left( \frac{u(x)+v(y)-c(x,y)}{\varepsilon} - 1 \right)d\mu(x)d\nu(y).
\end{equation}

\section{C-VAE formulation}

C-VAE generalizes InfoVAE via EOT, which will address the prior hole problem and relax the Gaussian assumption of the approximated posterior distribution.

\newterm{Framework.}  Assume all probability measures are absolutely continuous with respect to the Lebesgue measure or counting measure. i.e., $d\pi(x,y) = \pi(x,y)dxdy$. Consider data space $\gX$ and latent space $\gZ$ as the underlying metric space for EOT. Data distribution $P_{D}$ and the prior distribution of latent variable $P_Z$ from VAEs serve as marginals for the joint distribution. Let cost function $ c(x,z) $ to be negative log likelihood, i.e., $c(x,z) = - \log p_\theta(x|z)$. 
% Noting $p_\theta({x}|{z})$ is the likelihood function that the generative model objective tries to optimize, its negative logarithm is used as a cost function from the EOT perspective. It has a natural explanation as decoding cost such that if $x$ can be successfully decoded from $z$, then the cost is small. Otherwise, 
% the cost can be arbitrarily large.
Plugging in everything above into \eqref{Eq:EOT}, we get:
\begin{align}
OT_\varepsilon(P_{D},P_Z) 
= \min_{\pi(x,z)} &\int_{\gX\times \gZ} -\log p_\theta(x|z) \pi(x,z)dxdz \nonumber \\
&\quad+ \varepsilon \KL(\pi(x,z) \Vert p(z) p_{D}(x)). 
\end{align}

Decompose the joint distribution into marginal and conditional distribution, i.e., $\pi(x,z) = q(z|x)p_{D}(x)$. OT requires the joint distribution to satisfy the other marginal distribution also, i.e.,  $p(z) = q(z) \stackrel{{\mathsf{def}}}{=}\int q(z|x)p_{D}(x)dx$. Relaxing this hard constraint with a KL divergence and minimizing the transport loss with respect to the cost function, we get:
\begin{align}
OT_\varepsilon(P_{D},P_Z) 
&= \min_{\pi(x,z)} \int_{\gX\times \gZ}-\log p_\theta(x|z) \pi(x,z)dxdz \nonumber \\
&\qquad\qquad + \varepsilon \KL(\pi(x,z) \Vert p(z)p_{D}(x))\nonumber\\
&\qquad\qquad + \varsigma\KL(q(z) \Vert p(z)) \nonumber \\
&= \min_{q(z|x)} \E_{p_{D}(x)}\E_{q(z|x)}[-\log p_{\theta}(x|z)] \nonumber\\
&\qquad\qquad + \varepsilon \E_{p_{D}(x)}[D_{KL}(q_(z|x)\Vert p(z))]\nonumber\\
&\qquad\qquad + \varsigma\KL(q(z) \Vert p(z)),\label{Eq:otae}
\end{align}

where $\varsigma >0$ is the weight for the penalty term. Let $\varepsilon = 1-\alpha$ and $\varsigma = \alpha + \lambda -1$, \eqref{Eq:otae} coincides with \eqref{Eq:infovae} except the optimization is only about the posterior $q(z|x)$. With $p_\theta({x}|{z})$ as a neural network parametrized decoder, optimizing $\theta$ by SGD, we arrive at the final objective:
\begin{align}
&\min_{p_{\theta}(x|z)} \min_{q(z|x)} \E_{p_{D}(x)}\E_{q(z|x)}[-\log p_{\theta}(x|z)] \nonumber\\
&\quad + \varepsilon \E_{p_{D}(x)}[D_{KL}(q(z|x)\Vert p(z))] + \varsigma\KL(q(z) \Vert p(z)).
\end{align}

% Noting $p_\theta({x}|{z})$ is the likelihood function that the generative model objective tries to optimize, its negative logarithm is used as a cost function from the EOT perspective. It has a natural explanation as decoding cost such that if $x$ can be successfully decoded from $z$, then the cost is small. Otherwise, 
% the cost can be arbitrarily large.

Noting that $p_\theta({x}|{z})$ is the likelihood function in VAEs, which has a natural explanation as decoding cost such that if $x$ can be successfully decoded from $z$, then the cost is small. Otherwise, the cost is large. Cost function optimization is problematic in the usual OT setting because the cost function can be arbitrarily small for any pair of $x$ and $z$. However, this is not the case for our model as the likelihood is assumed to be Gaussian or Bernoulli. 
\begin{proposition} Let $p_\theta (x|z)$ be selected from the family of all parametric Gaussian distributions, i.e.,  $p_{\theta}(x|z) = \gN(x| \mu_\theta(z), \mathrm{diag}(\sigma_\theta^2(z)))$. Then for any $z \in \gZ$, $c(\cdot , z) = -\log p_\theta (\cdot |z)$ is a positive quadratic function with minimum $\sum_{i=1}^{d_x} \log \sqrt{2\pi \sigma^2_{\theta,i}(z)}$ achieved at $ x = \mu_\theta(z)$.

\end{proposition}
The proof is a direct calculation of the negative log density function of Gaussians. It shows that given any $z$, there can be only one minimum at the mean of the distribution. If we want to make all costs equally small, then the minimum $\sum_{i=1}^{d_x} \log \sqrt{2\pi \sigma^2_{\theta,i}(z)}$ will go up as $ \sigma^2_{\theta,i}(z)$ will increase. Similar results can be derived for Bernoulli likelihood. 

\newterm{EOT acts as variational inference.}
In VAEs, amortized variational inference  requires approximated posteriors $q_\phi(z|x)$ to be multivariate gaussian distributions which are modeled by neural networks with parameters $\phi$. Furthermore, the covariance matrices of Gaussians are often assumed to be diagonal. Normalizing flow enriches the distributions via composing with invertible transformations. But it is still within a restricted family of distributions. This limits the expressiveness of the variational distributions and hurts the accuracy of the inference. In the OT setting, we can solve for the approximated posterior $q(z|x)$ or more generally joint distribution $\pi(z,x)$ using optimal transport solvers such as the Sinkhorn algorithm in the discrete setting, stochastic gradients method of Dual or Semi-dual in the continuous setting, and unbalanced Sinkhorn for relaxed marginals problem without assuming the form of approximate distributions.

\newterm{EOT explanations of VAEs.} One of the well-known drawbacks of VAEs models is the blurry samples. People originally attributed this issue on the maximum likelihood objective that penalizes differently when $p_\theta(x) > p_D(x)$ and when $p_\theta(x) < p_D(x)$. \citet{Zhao2017TowardsDU} argued that blurriness is not merely because of the objective but the VAE approximation of the maximum likelihood objective, whereas \citet{Cai2017MultiStageVA} argued the $L_2$ distance used in the objective caused the fuzziness in samples. We propose to explain the fuzziness in samples through EOT. The encoder's and decoder's parameter learning in VAEs correspond to the coupling and cost function learning in EOT. The relative entropy term forces the coupling to spread over all possible locations, resulting in a plan $\pi(x,z)$ that has a non-zero density between different $x_i$'s and the same $z$. Then if we try to optimize the cost function to further reduce the transport cost, based on the proposition we state above, we know that the decoder will send $z$ to the weighted average of all $x_i$'s. 

Another common problem of VAEs is the posterior collapse phenomenon, which refers to the situation when $q_\phi(z|x) = p(z)$ for any $x$. One commonly accepted reason is that the decoder is too flexible. It could ignore the latent representations but achieve a good enough likelihood estimation, i.e., $p_\theta(x|z) = p_D(x)$. This is also easy to explain in EOT as the cost function is no longer a function of $z$. The optimal plan only depends on the relative entropy term which will achieve minimum when the plan is an independent coupling.

% For the prior hole, it's just because the limited capacity of the model and no explicit constraints on marginals.

\newterm{Matching the aggregate posterior with prior.} In the following sections, we will assume $\varsigma$ = $+\infty $ which forces aggregated posterior to equal prior distribution (hard constraint). We claim this is not an unreasonable choice because maximizing VAE objective with respect to the prior distribution is equivalent to matching the prior with aggregate posterior, and if the model has learned the data distribution, i.e. $p_\theta(x) = p_D(x)$ and approximated posterior capture the true posterior, i.e., $q_\phi(z|x) = p(z|x)$, then  $q_{\phi}(z) = \int q_{\phi}(z|x)p_D(x)dx = \int p(z|x)p_\theta(x)dx$ should equal to $p(z)$.

Unlike the primal OT problem which is a constrained optimization problem that is known to be difficult to solve, the Dual and Semi-Dual formulation is unconstrained where SGD methods apply naturally:
\begin{align}\small
\min_\theta \max_{\substack{u \in \gC(\gX) \\ v \in \gC(\gZ)}} \E_{p(z) \otimes p_D(x)}\left[  u(x) + v(z) 
- \varepsilon e^{\frac{u(x)+v(z)-c_{\theta}(x,z)}{\varepsilon}} \right] 
\end{align}
and
\begin{equation}
\min_\theta \max_{u \in \gC(\gX)} \E_{p(z)} \left[\int_{\gX}u(x)p_D(x)dx + u^{c_{\theta},\varepsilon}(z) \right].
\end{equation}

We keep $c_\theta(x,z) = - \log p_\theta(x|z)$ for readability. Since $p_D(x)\approx \frac{1}{n}\sum_{i=1}^n \delta_{x_i}$ where $\delta_{x}$ is Dirac-delta function, inner optimization is actually a finite-dimensional concave maximization problem:
\begin{equation}
   \min_{{\theta}} \max_{\mathbf{u} \in \R^n} \mathcal{L}_\mathbf{C-VAE},
\end{equation}
where 
\begin{equation*} 
\mathcal{L}_\mathbf{C-VAE} = \E_{p(z)} \left[\sum_{i=1}^n \frac{1}{n} \mathbf{u}_i   -\varepsilon \log \sum_{i=1}^n \frac{1}{n} e^{\frac{\mathbf{u}_i - c_\theta(x_i,z)}{\varepsilon}} \right].
\end{equation*}
% \begin{equation}
% \min_{{\theta}} \max_{\mathbf{u} \in \R^n} \E_{p(z)} [\sum_{i=1}^n \frac{1}{n} \mathbf{u}_i   -\varepsilon \log \sum_{i=1}^n \frac{1}{n} \exp \left({\frac{\mathbf{u}_i - c_\theta(x,z)}{\varepsilon}}\right)],
% \end{equation}

With the primal-dual relationship, we can recover the optimal plan:
\begin{equation} \small
\pi(x_i,z) = \exp \left( \frac{\mathbf{u}_i + u^{c_{\theta},\varepsilon}(z)-c_\theta(x_i,z)}{\varepsilon} - 1 \right)p(z)p_{D}(x_i).   
\end{equation}

Substitute $\pi$ with $ q(z|x)p_{D}(x_i)$, we can get the expression of posterior
 :
\begin{equation} \small
q(z|x_i) = \exp \left( \frac{\mathbf{u}_i + u^{c_{\theta},\varepsilon}(z)-c_\theta(x_i,z)}{\varepsilon} - 1 \right)p(z).  
\end{equation}

\newterm{Flexible choice of prior distribution.}
Our model can work with any prior distribution from which we can sample. There are no differences in training between models with different priors. The data distribution $P_D$ is always discrete as it is approximated by Dirac measures of samples in the training set empirically. If we have a discrete prior over the latent variables, i.e., categorical distribution  $P_Z = Cat(K;\mathbf{p})$,  we will arrive at a discrete EOT problem that can be solved by Sinkhorn algorithm efficiently. 

We will arrive at semi-discrete EOT problems if we have continuous marginal over latent variables like $ P_Z  = \gN(\mathbf{z};\mathbf{0},\mathbf{I})$ or $ P_Z  = Dir(\mathbf{z};\mathbf{K},\mathbf{\alpha})$. Next, we will come up with two different strategies for learning the latent generative models. The main difference between these two strategies is  optimizing the primal or dual formulation of EOT with respect to the likelihood function.

\begin{algorithm}
\caption{Primal Strategy}\label{algo:primal}
\begin{algorithmic} 
\REQUIRE 
Input distributions $P_Z$, $P_D$; cost function or decoder network $p_\theta(x|z)$; sample size $n$; learning rate $\lambda_1$, $\lambda_2$; number of iterations $K$;\\
Semi-Dual: Dual variable $\mathbf{u}$ or $u_\phi$ if parametrized by neural network;\\
Dual: Dual variables $\mathbf{u}$, $\mathbf{v}$ or $u_\phi$, $v_\psi$ if parametrized by neural network;
\ENSURE $\theta, \phi$ (and $\psi$ if dual formulation)
\REPEAT
\FOR{$k = 1,2,...K$}
    \item Sample $(z_j)_{j=1}^n \sim P_{Z}$;
    \STATE $\mathcal{L}_{\mathbf{u}} \leftarrow \frac{1}{n}\sum_{j=1}^n[\sum_{i=1}^m \frac{1}{m} \mathbf{u}_i  -$
    \STATE $\qquad\qquad\quad\varepsilon \log \sum_{i=1}^m \frac{1}{m} \exp \left({\frac{\mathbf{u}_i - c_\theta(x_i,z_j)}{\varepsilon}}\right)]$
    \item Update $\mathbf{u}$ using $\frac{\partial\mathcal{L}_{\mathbf{u}}}{\partial\mathbf{u}}$ and $\lambda_1$ to maximize $\mathcal{L}_{\mathbf{u}}$;
\ENDFOR

\item Sample $(z_j)_{j=1}^n \sim Q_{Z|X}$;
\STATE $\mathcal{L}_{\mathbf{\theta}} \leftarrow \frac{1}{m}\sum_{i=1}^m[\frac{1}{n}\sum_{j=1}^n  -\log p_{\theta}(x_i|z_j)]$
\item Update $\theta$ using $\frac{\partial\mathcal{L}_{\theta}}{\partial\theta}$ and $\lambda_2$ to minimize $\mathcal{L}_{\theta}$;
\UNTIL converged or reach the max number of epochs

\end{algorithmic}
\end{algorithm}

\subsection{Optimizing $q(z|x$) with dual but $p_\theta(x|z)$ with primal}
Building on the foundation from the above section, we now define the first training strategy that is established on the primal formulation of EOT. We call this Primal Strategy (see Algorithm~\ref{algo:primal}):
\begin{enumerate}[noitemsep,topsep=0pt]
  \item Solve for the optimal $q(z|x)$ in dual EOT with a fixed log likelihood $\log p_\theta(x|z) $ as cost function.
  \item Optimize the primal EOT objective with respect to likelihood function $p_\theta(x|z) $ using posterior $q(z|x)$ got from the first step.
  \item Repeat these two steps until convergence or desired number of steps.
\end{enumerate}

Step one is to find the optimal plan or optimal conditional plan given the marginals. In this step, we will use dual or semi-dual formulation depending on the context. By primal-dual relationship, we can get $q(z|x)$ from dual variables. Then, we are solving for the likelihood function which needs to evaluate the expectation with respect to $q(z|x)$. This is particularly difficult as the integral does not have a closed form. Monte Carlo estimate is also hard because samples from a nontrivial $q(z|x)$ are difficult. Fortunately, the expression of $q(z|x)$ derived above could be seen as a weighted prior which is exactly what we need in importance sampling. The computationally expensive parts of the weight can be cached during the optimization of the optimal plan. Calculating importance weight will not cost much to generate samples from the posterior.

%\newterm{Importance Sampling.} An interesting and useful side product of the above derivation is that the posterior $q(z|x)$ is perfectly suitable for importance sampling. The posterior could be seen as a weighted prior whose weight will be evaluated during searching for the optimal plan. In other words, calculating importance weight will not cost anything to generate samples from the posterior.

\begin{algorithm}
\caption{Dual Strategy}\label{algo:dual}
\begin{algorithmic} 
\REQUIRE 
Input distributions $P_Z$, $P_D$; cost function or decoder network$p_\theta(x|z)$; sample size $n$; learning rate $\lambda_1$, $\lambda_2$; number of iterations $K$;\\
Semi-Dual: Dual variable $\mathbf{u}$ or $u_\phi$ if parametrized by neural network;\\
Dual: Dual variables $\mathbf{u}$, $\mathbf{v}$ or $u_\phi$, $v_\psi$ if parametrized by neural network;
\ENSURE $\theta, \phi$ (and $\psi$ if dual formulation)
\REPEAT
\FOR{$k = 1,2,...K$}
    \item Sample $(z_j)_{j=1}^n \sim P_Z$;
    \STATE $\mathcal{L}_{\mathbf{u}} \leftarrow \frac{1}{n}\sum_{j=1}^n[\sum_{i=1}^m \frac{1}{m} \mathbf{u}_i -$   
    \STATE $\qquad\qquad\quad \varepsilon \log \sum_{i=1}^m \frac{1}{m} \exp \left({\frac{\mathbf{u}_i - c_\theta(x_i,z_j)}{\varepsilon}}\right)]$
    \item Update $\mathbf{u}$ using $\frac{\partial\mathcal{L}_{\mathbf{u}}}{\partial\mathbf{u}}$ and $\lambda_1$ to maximize $\mathcal{L}_{\mathbf{u}}$;
\ENDFOR

\item Sample $(z_j)_{j=1}^n \sim P_Z$;
\STATE $\mathcal{L}_{\mathbf{\theta}} \leftarrow \frac{1}{n}\sum_{j=1}^n[\sum_{i=1}^m \frac{1}{m} \mathbf{u}_i -$   
\STATE $\qquad\qquad\quad \varepsilon \log \sum_{i=1}^m \frac{1}{m} \exp \left({\frac{\mathbf{u}_i - c_\theta(x_i,z_j)}{\varepsilon}}\right)]$
% \item Update $\mathbf{u}$ using $\frac{\partial\mathcal{L}_{\mathbf{u}}}{\partial\mathbf{u}}$ and $\lambda_2$ to maximize $\mathcal{L}_{\mathbf{u}}$;
\item Update $\theta$ using $\frac{\partial\mathcal{L}_{\theta}}{\partial\theta}$ and $\lambda_2$ to minimize $\mathcal{L}_{\theta}$;
\UNTIL converged or reach the max number of epochs

\end{algorithmic}
\end{algorithm}

\subsection{Optimizing $q(z|x$) with dual and $p_\theta(x|z)$ with dual}

An alternative training strategy is to calculate $q(z|x)$ by the primal-dual relationship, it is possible to optimize dual objective with $p_{\theta}(x|z)$ directly (see Algorithm~\ref{algo:dual}). The relationship between optimal coupling and dual variables only holds at optimum. When we update only for a fixed number of steps instead of convergence, the coupling can be very biased. Direct optimizing dual can improve this problem.
Based on \citet{genevay16}, we could derive the gradients with respect to dual variable and cost function analytically.

\begin{proposition}
When $\epsilon >0$, the gradient of $ \mathcal{L}_\mathbf{C-VAE}$ with respect to $c_\theta$ is given by $$\left(\frac{\partial\mathcal{L}_\mathbf{C-VAE}}{\partial c_\theta} \right)_j = \E_{p(z)} \left[\frac{ \exp \left({\frac{\mathbf{u}_j - c_\theta(x_j,z)}{\varepsilon}}\right)] }{\sum_{i=1}^n \exp \left({\frac{\mathbf{u}_i - c_\theta(x_i,z)}{\varepsilon}}\right)] }\right] $$

% and $(\frac{\partial\mathcal{L}_\mathbf{C-VAE}}{\partial c_\theta})_j \in (0,1) $ for all $j$.
\end{proposition}
Proof is trivial. For details we refer to \citet{genevay16} where gradients of dual variables are given.
By the chain rule, the decoder can be trained through back-propagation.

% A key advantage of treating posterior inference as an EOT problem is that we don't need to assume isotropic Gaussian distributions for both prior distribution and posterior anymore. Expressive prior and approximated posterior are proved to be essential in modeling data distribution and learning useful representations. 

\section{Related work}
There are several examples of OT-based generative modeling in the past few years. The majority focus on classical optimal transport \citep{Arjovsky2017WassersteinGA,tomczak18a, Patrini2018SinkhornA, Seguy, Deshpande2018GenerativeMU, An2020AeOTAN, Rout2021GenerativeMW}. WGAN \citep{Arjovsky2017WassersteinGA} replaced the Jensen-Shannon divergence optimized in the original GAN framework with the Wasserstein-1 distance. \citet{Deshpande2018GenerativeMU} proposed to further modify the GAN by approximating Wasserstein-1 distance by sliced Wasserstein distance. People also tried to apply OT on auto-encoder models. WAE \citep{tomczak18a} aims to minimize the penalized Wasserstein distance between model distribution and target distribution. SAE \citep{Patrini2018SinkhornA} replaced the MMD or GAN penalty term in WAE with Sinkhorn divergence. Sinkhorn generative model \citep{genevay18a} minimizes the Sinkhorn divergence between data distribution and generative distribution in a mini-batch manner. LSOT \citep{Seguy}, AE-OT \citep{An2020AeOTAN} and OTM \citep{Rout2021GenerativeMW} are computing the OT maps instead of the plans in generative modeling. LSOT considers continuous OT with regularization. AE-OT solves the semi-discrete OT between a noise distribution and encoded data distribution captured by an autoencoder. OTM is trying to find the OT maps in observation space that is different from AE-OT which happens in latent space. 

Our methods are different from the above methods as we compute the EOT loss between two distributions supported on different spaces. We unify the training of the encoder in VAEs with the EOT problem and treat the decoder as a learnable cost function. The recast of VAE as EOT provides a new perspective of generative modeling and new explanations for problems that occurred  during training.

% WAE\footnote{WAE has ppenalty parameter equaling 1 for Mixture of Gaussians and 2 for MNIST.} and InfoVAE\footnote{InfoVAE has hyperparameters $\alpha=0$, $\lambda=2$ for Mixture of Gaussians and $\alpha=0$, $\lambda=3$ for MNIST.}

\section{Experiments}
In this section, we explore various properties of C-VAE on a selection of synthetic and real datasets. And We also compare it with other autoencoder models including VAE, VAE-NF\footnote{VAE-NF means VAE with normalizing flow. In particular, We used 10 layers of planar flow.}, WAE\footnote{WAE has penalty parameter equaling 1 for Mixture of Gaussians and 2 for MNIST.} and InfoVAE\footnote{InfoVAE has hyperparameters $\alpha=0$, $\lambda=2$ for Mixture of Gaussians and $\alpha=0$, $\lambda=3$ for MNIST.}. All of the models are trained with the exact same architecture (if they have the same components) across all experiments. We use Dual strategy for C-VAE training in all experiments.
\begin{figure*}[ht]
  \centering
  % \rotatebox[origin=c]{90}{\bfseries Model 1\strut}
  
  \raisebox{0.5in}{\rotatebox[origin=t]{90}{VAE}} 
  \subfigure{\includegraphics[width=0.17\textwidth]{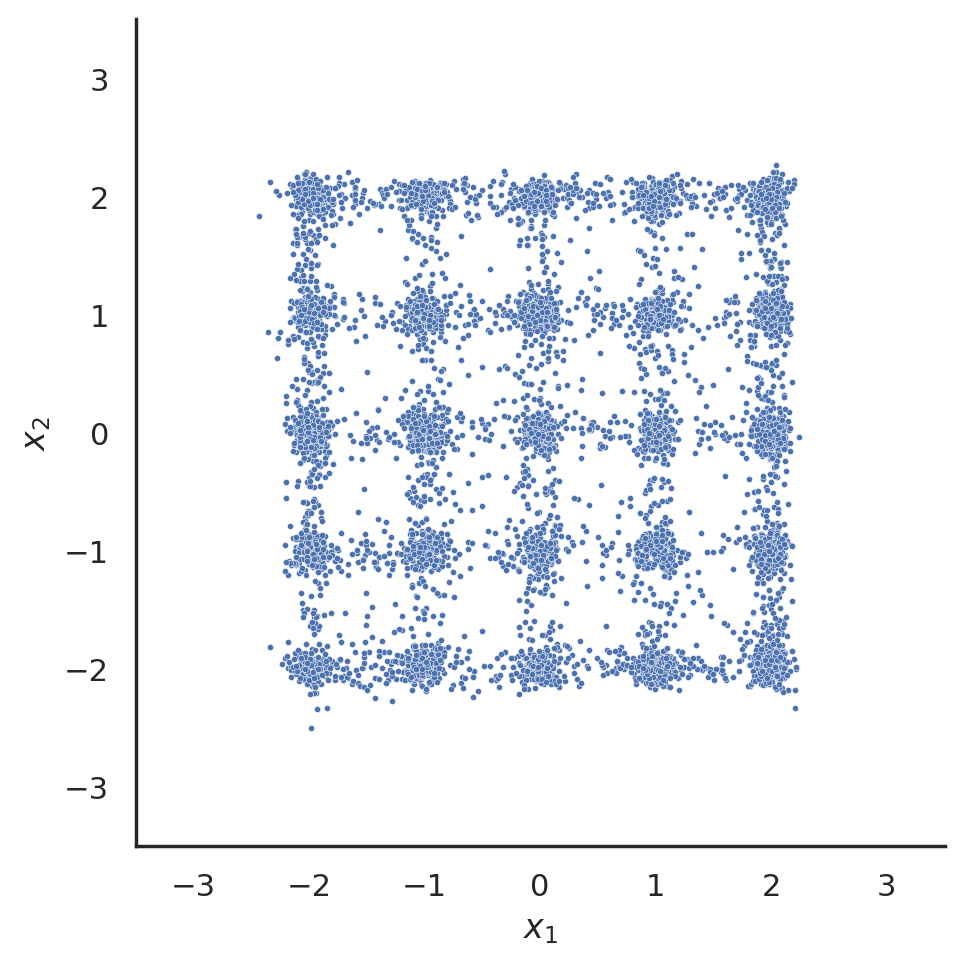}}
  \subfigure{\includegraphics[width=0.17\textwidth]{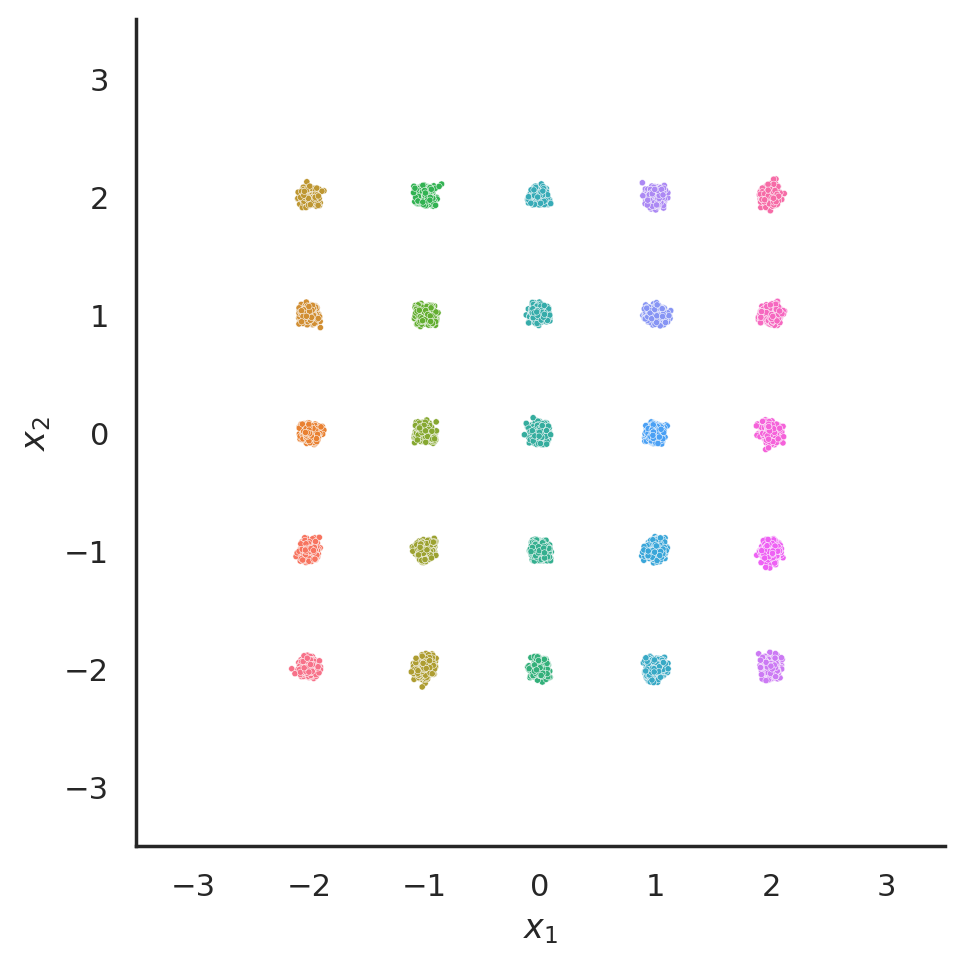}}
  \subfigure{\includegraphics[width=0.17\textwidth]{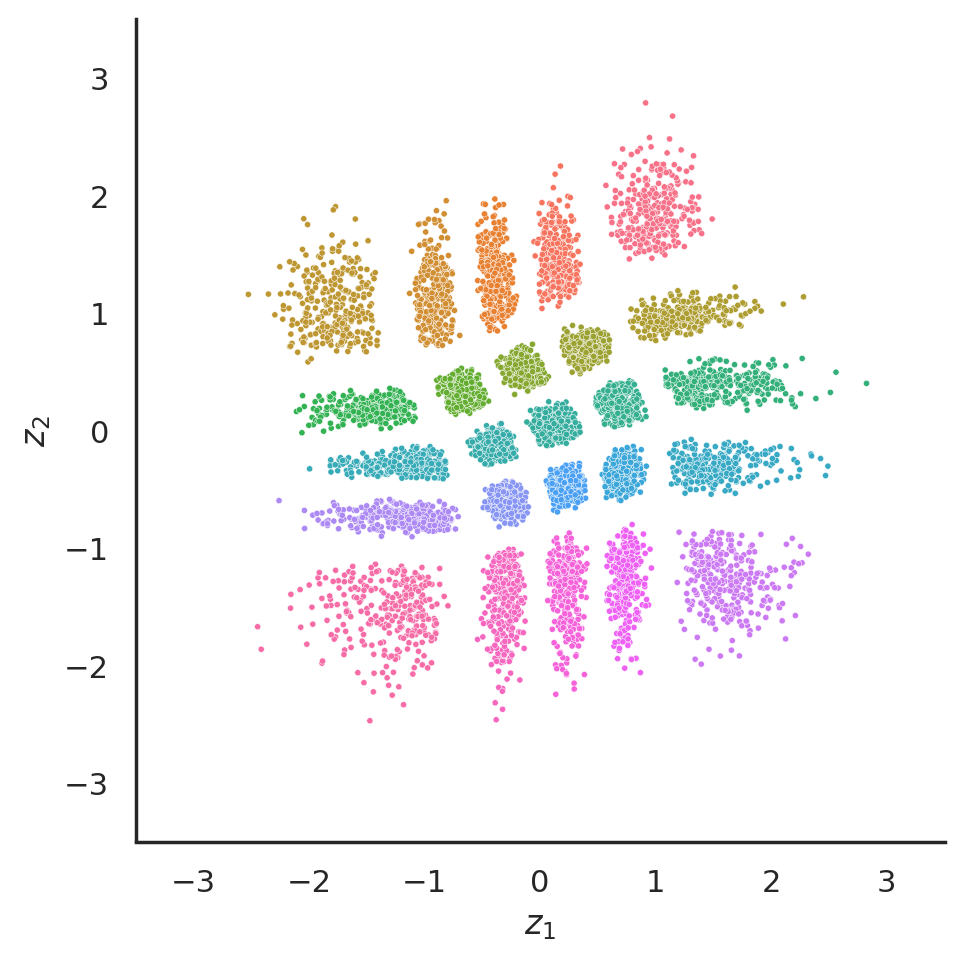}}
  \subfigure{\includegraphics[width=0.17\textwidth]{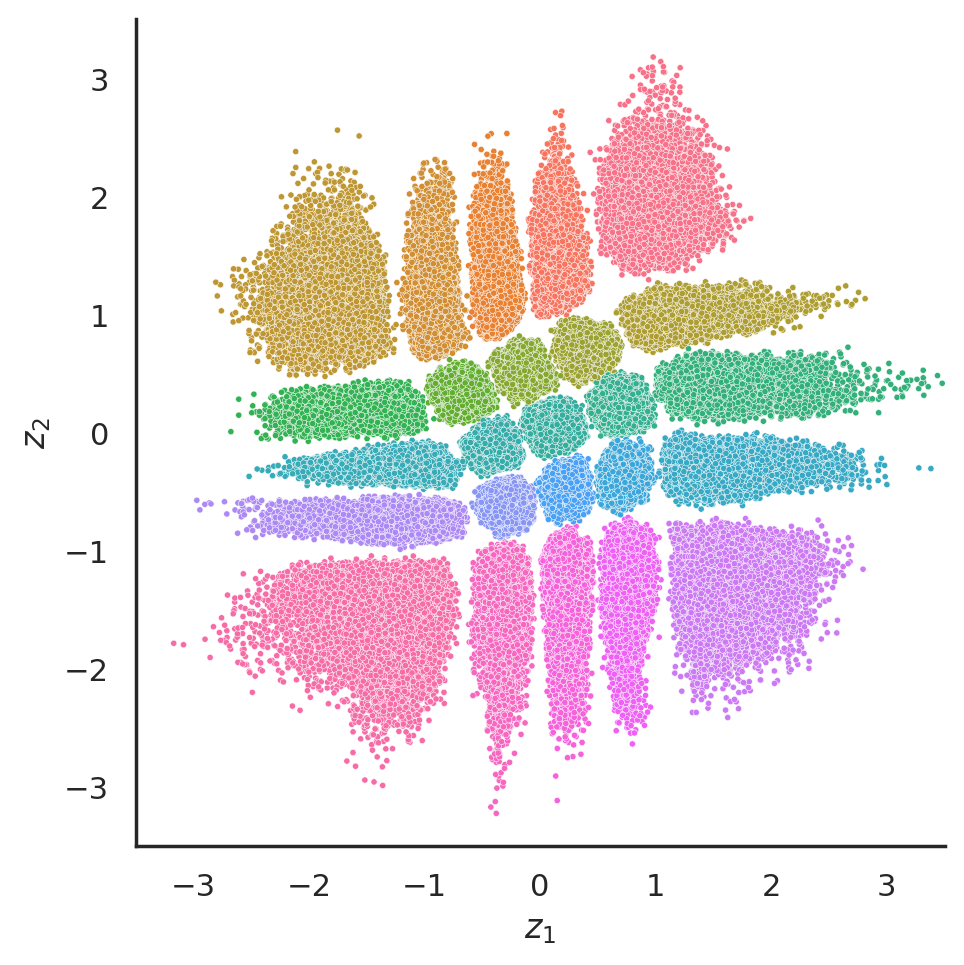}}
  \subfigure{\includegraphics[width=0.17\textwidth]{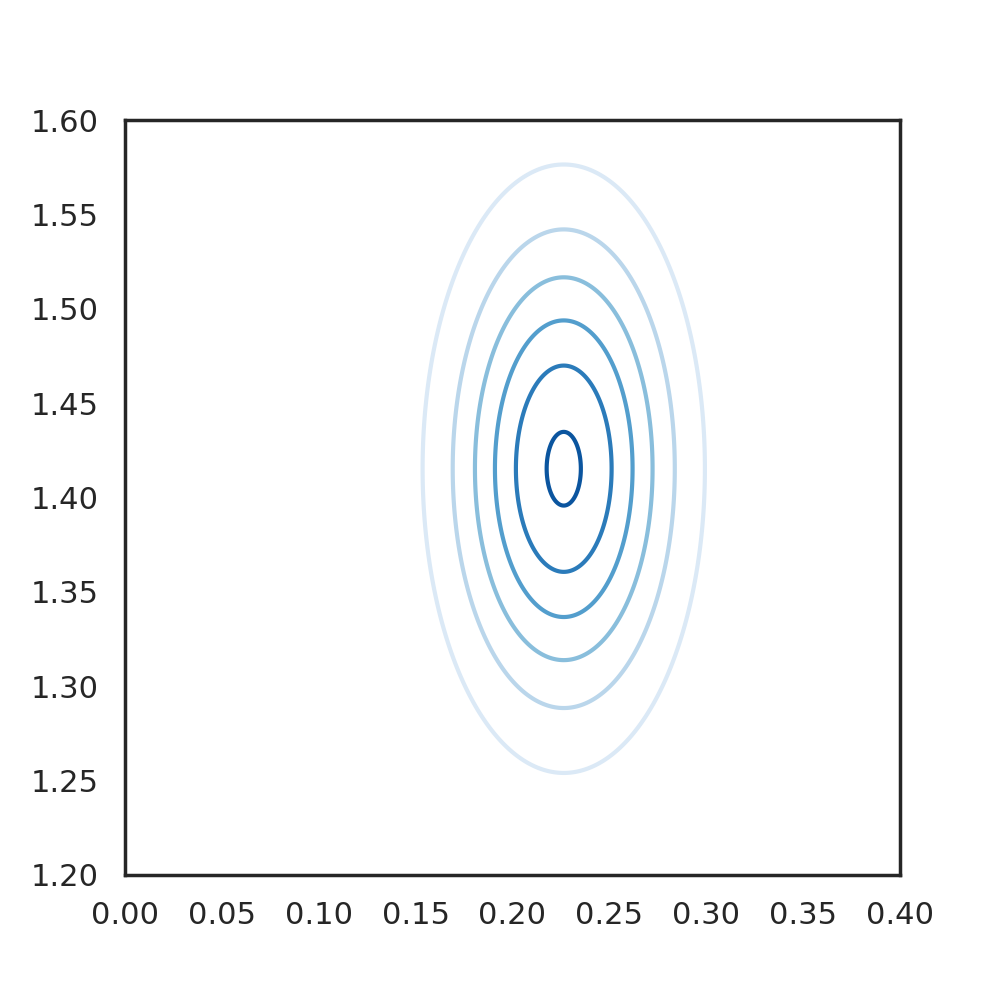}}

  \vspace{-5mm}
  \raisebox{0.5in}{\rotatebox[origin=t]{90}{VAE-NF}} 
  \subfigure{\includegraphics[width=0.17\textwidth]{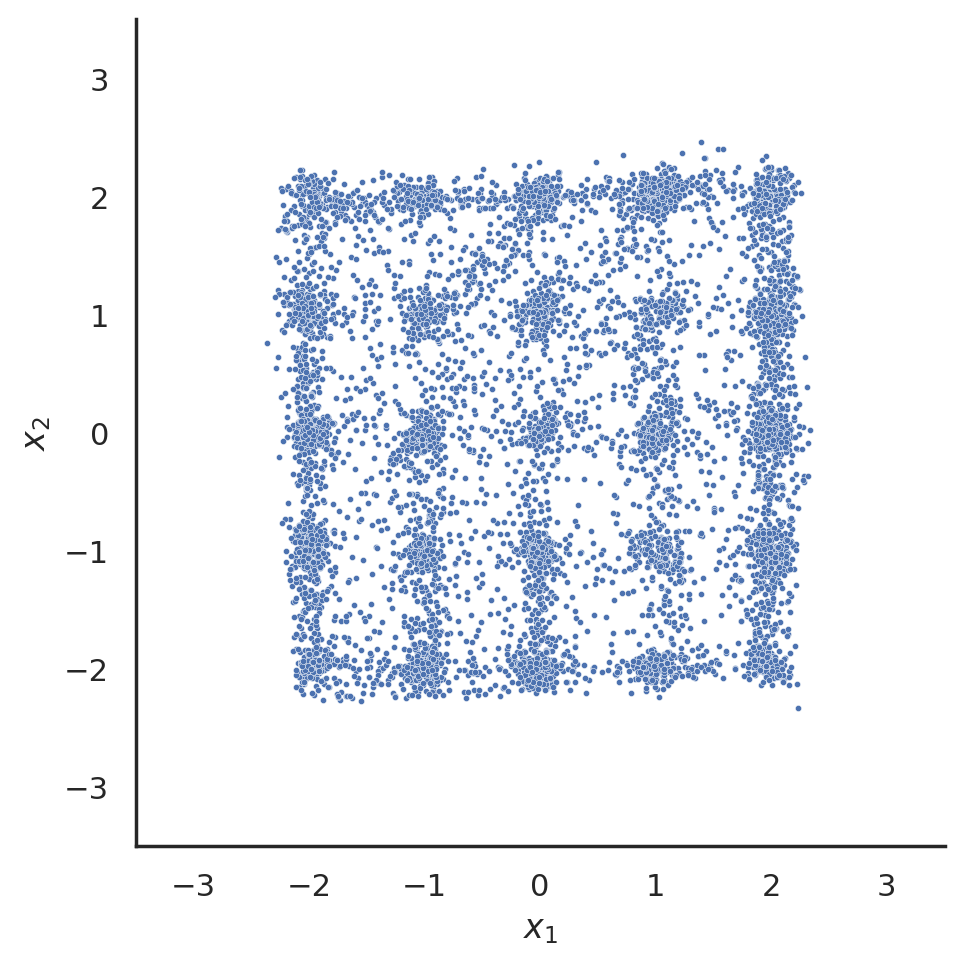 }}
  \subfigure{\includegraphics[width=0.17\textwidth]{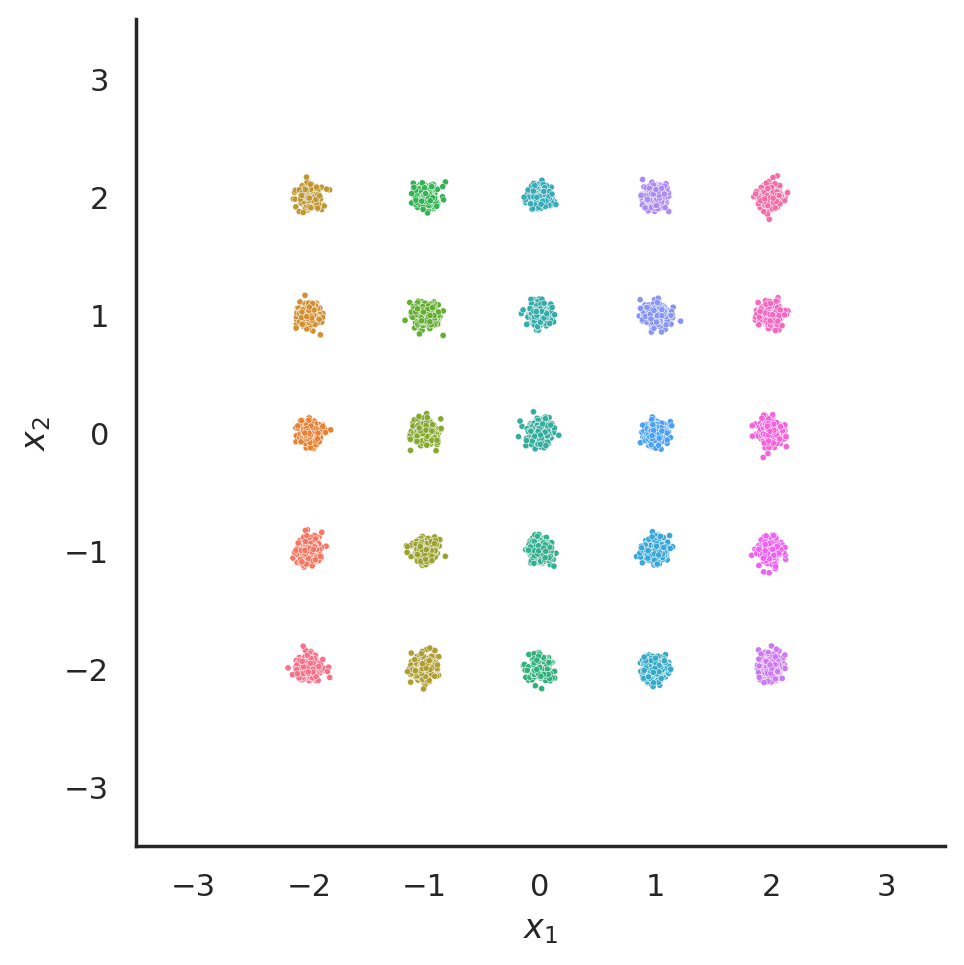}}
  \subfigure{\includegraphics[width=0.17\textwidth]{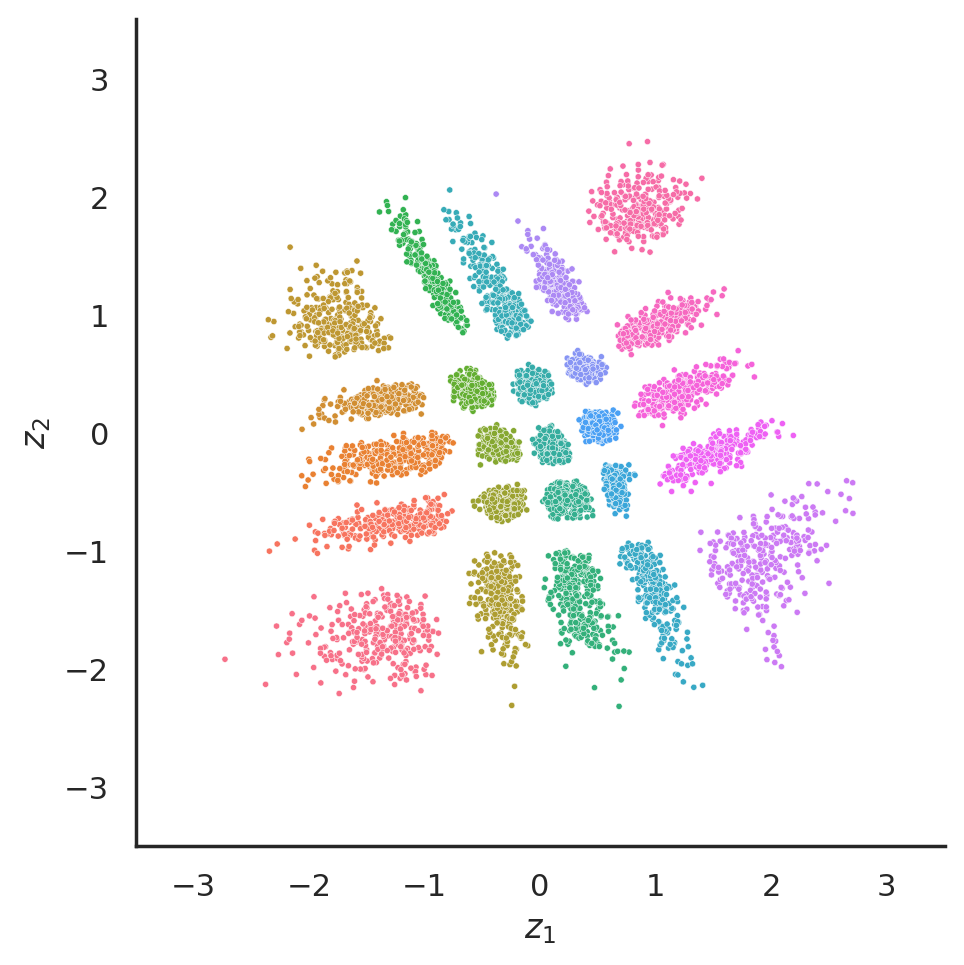}}
  \subfigure{\includegraphics[width=0.17\textwidth]{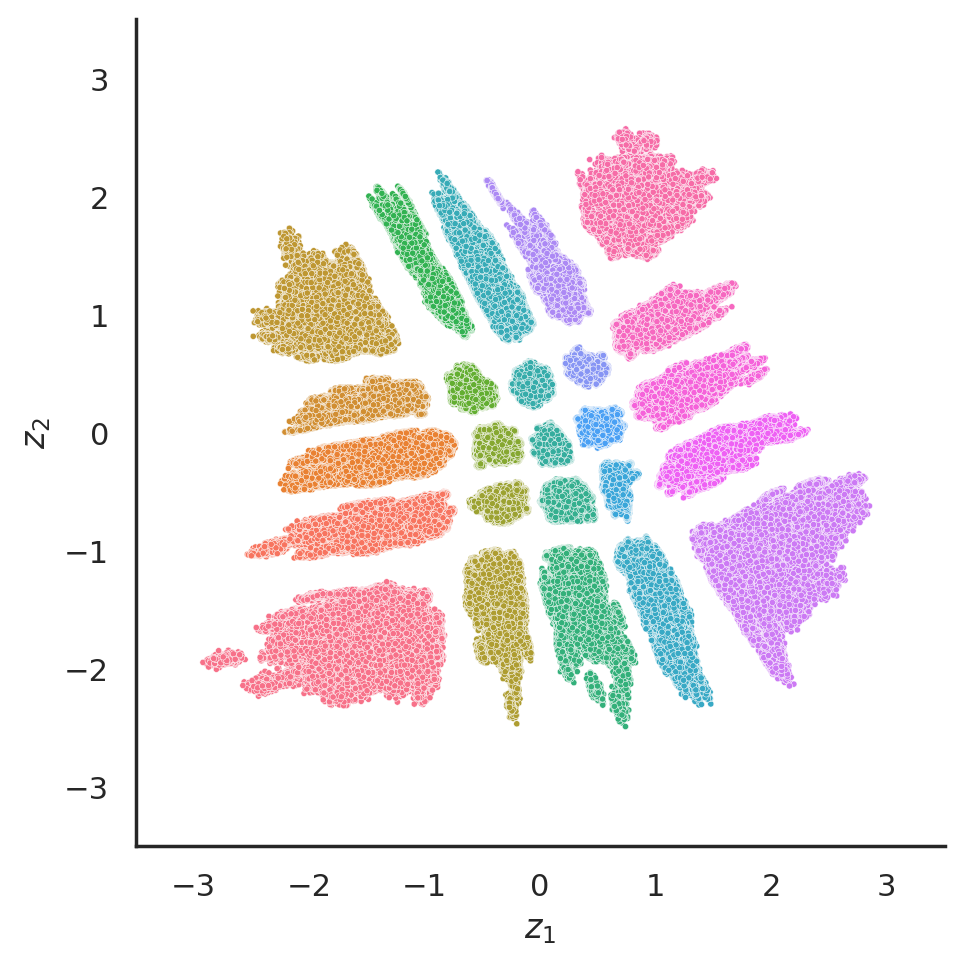}}
  \subfigure{\includegraphics[width=0.17\textwidth]{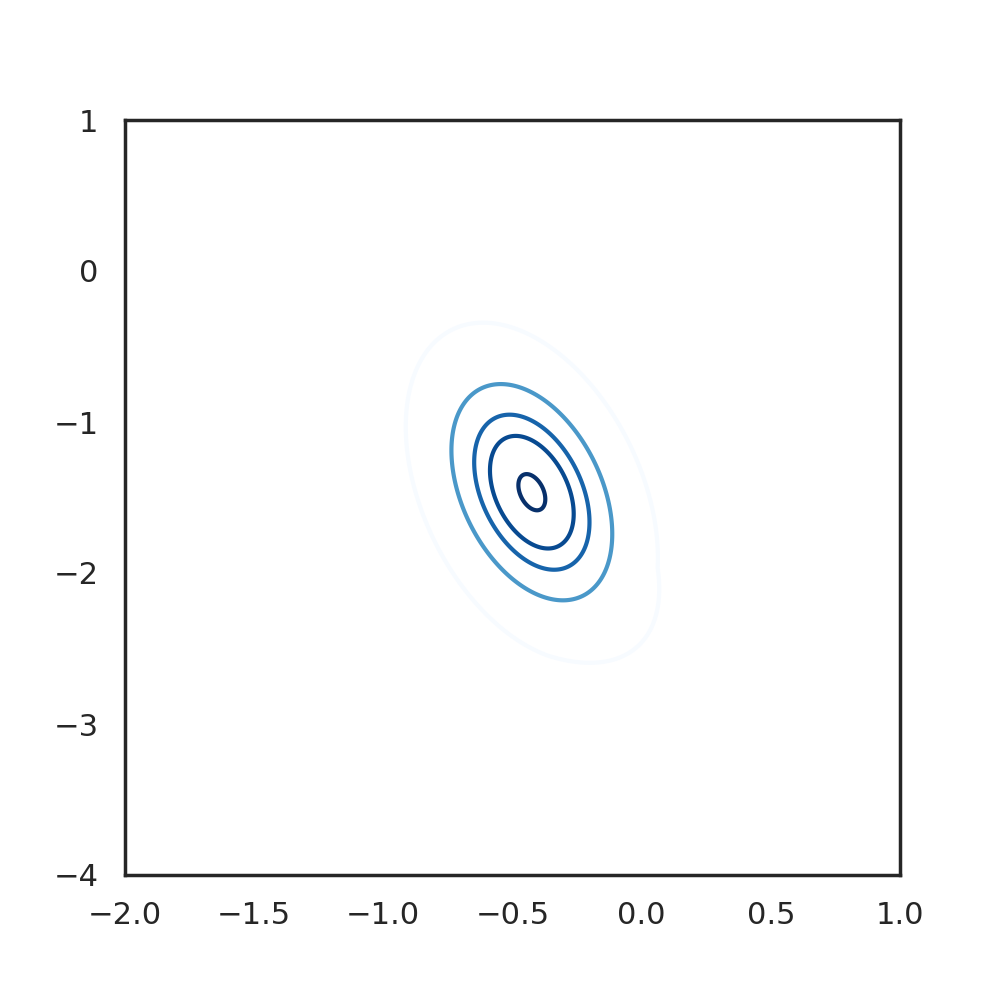}}

  \vspace{-5mm}
  \raisebox{0.5in}{\rotatebox[origin=t]{90}{WAE}} 
  \subfigure{\includegraphics[width=0.17\textwidth]{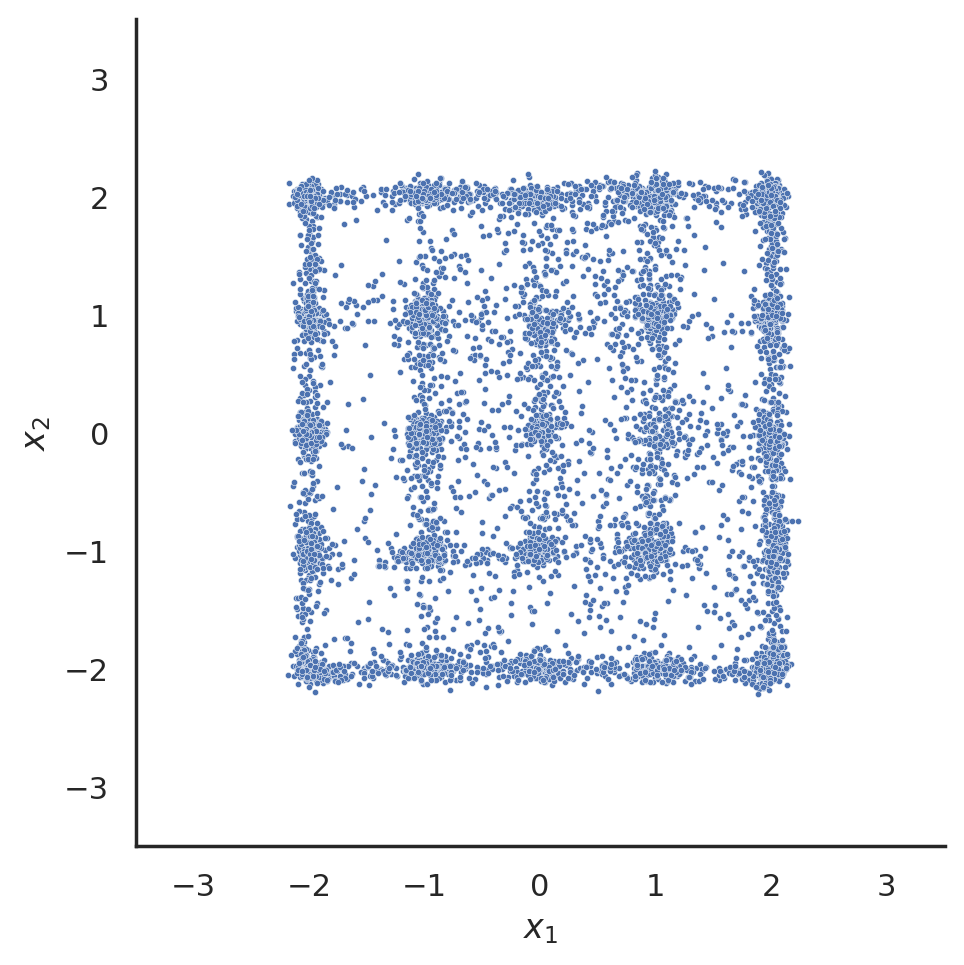}}
  \subfigure{\includegraphics[width=0.17\textwidth]{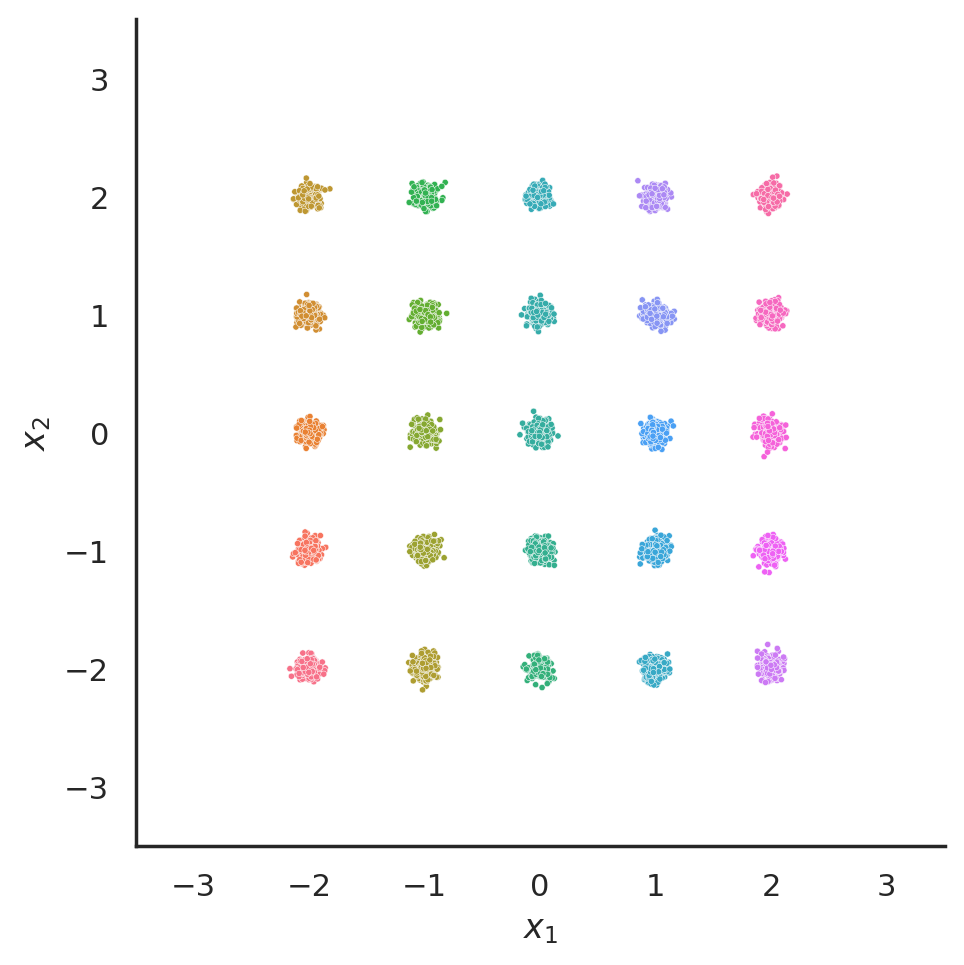}}
  \subfigure{\includegraphics[width=0.17\textwidth]{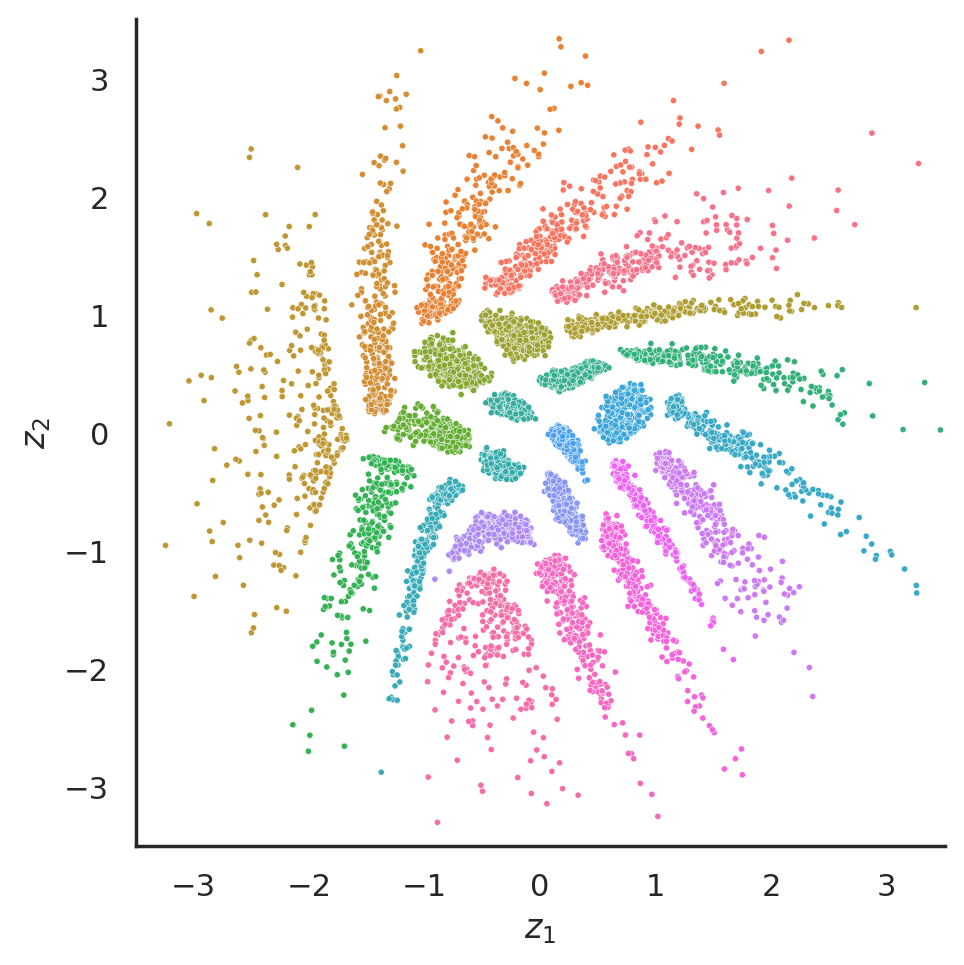}}
  \subfigure{\includegraphics[width=0.17\textwidth]{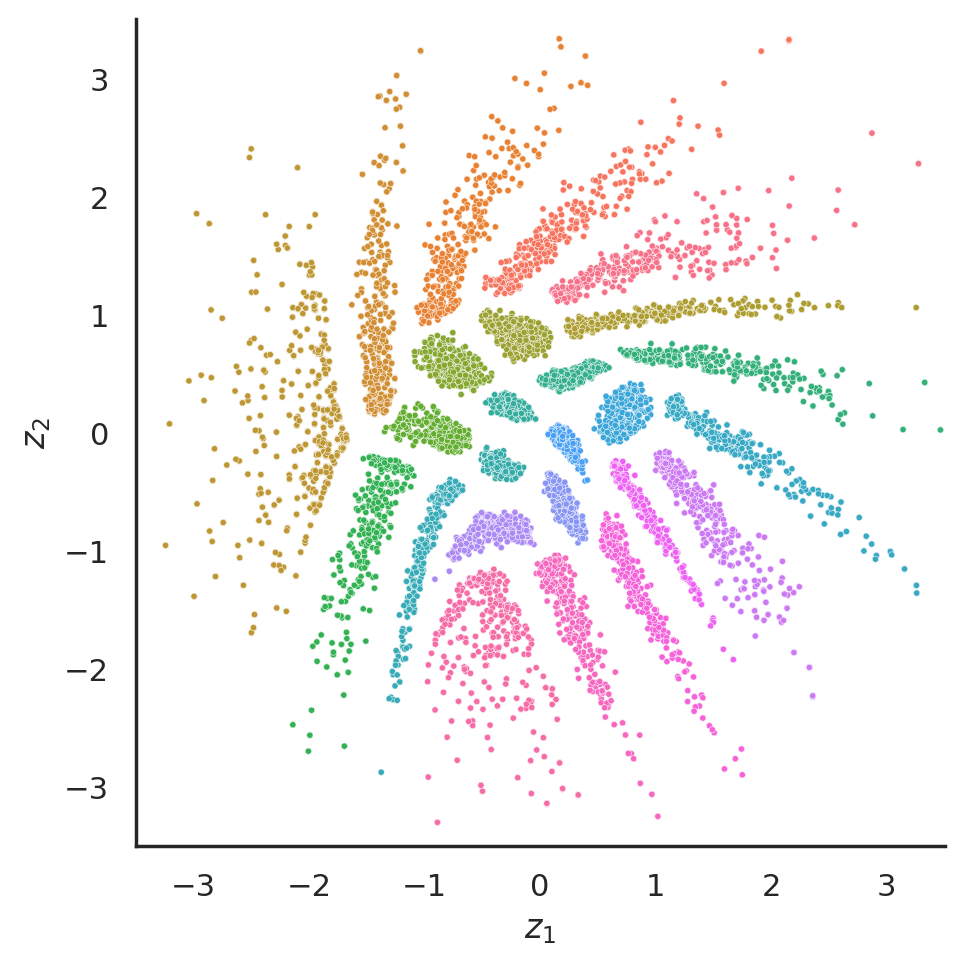}}
  \subfigure{\includegraphics[width=0.17\textwidth]{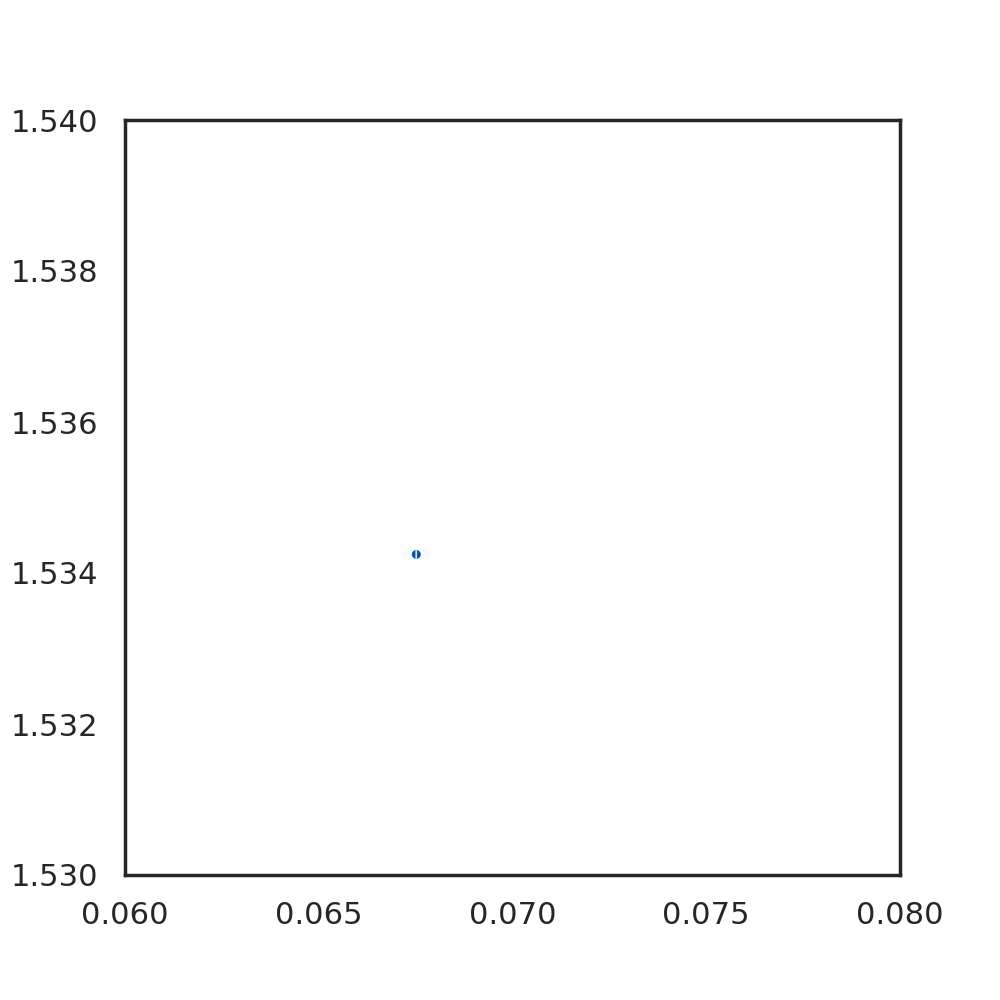}}

  \vspace{-5mm}
  \raisebox{0.5in}{\rotatebox[origin=t]{90}{InfoVAE}} 
  \subfigure{\includegraphics[width=0.17\textwidth]{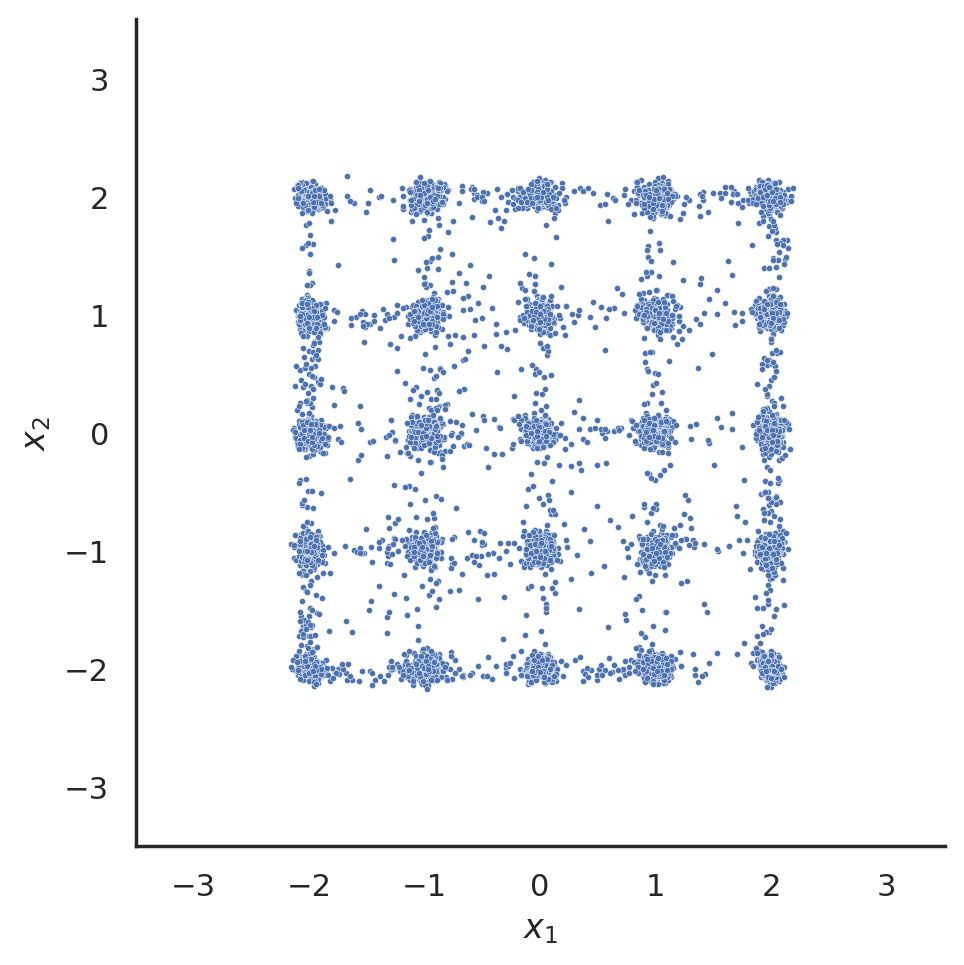}}
  \subfigure{\includegraphics[width=0.17\textwidth]{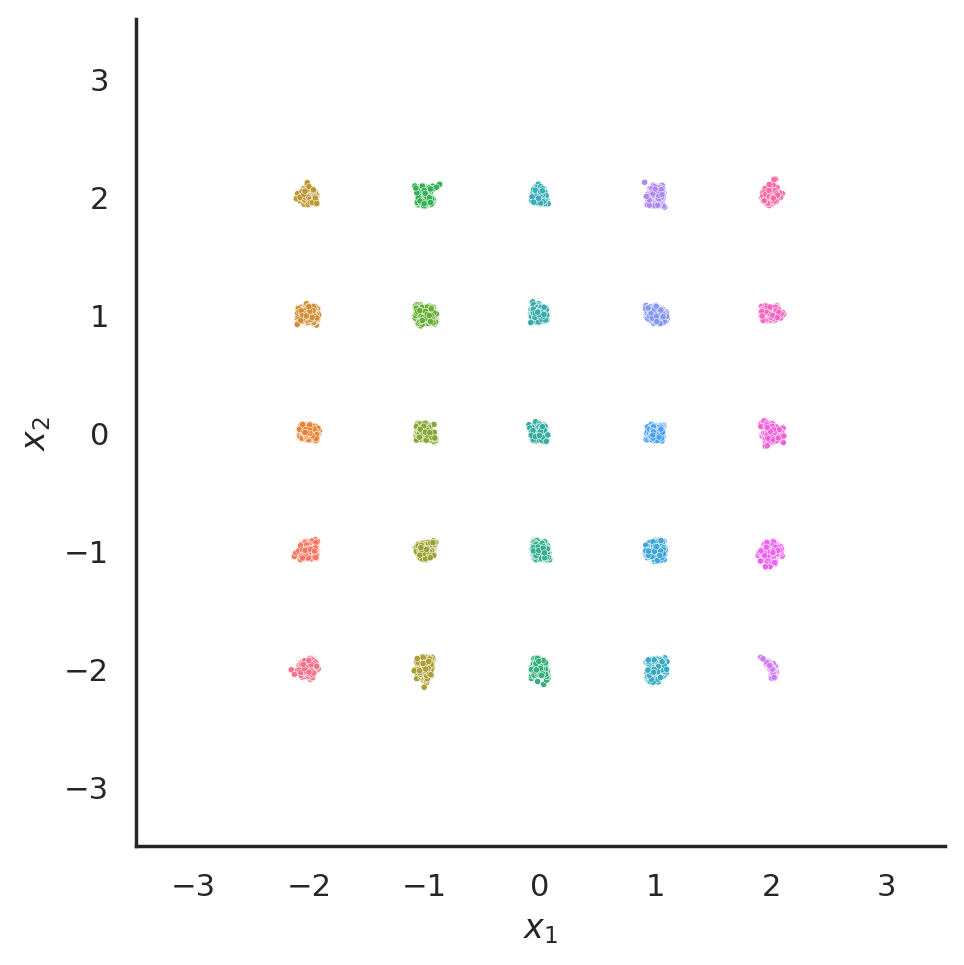}}
  \subfigure{\includegraphics[width=0.17\textwidth]{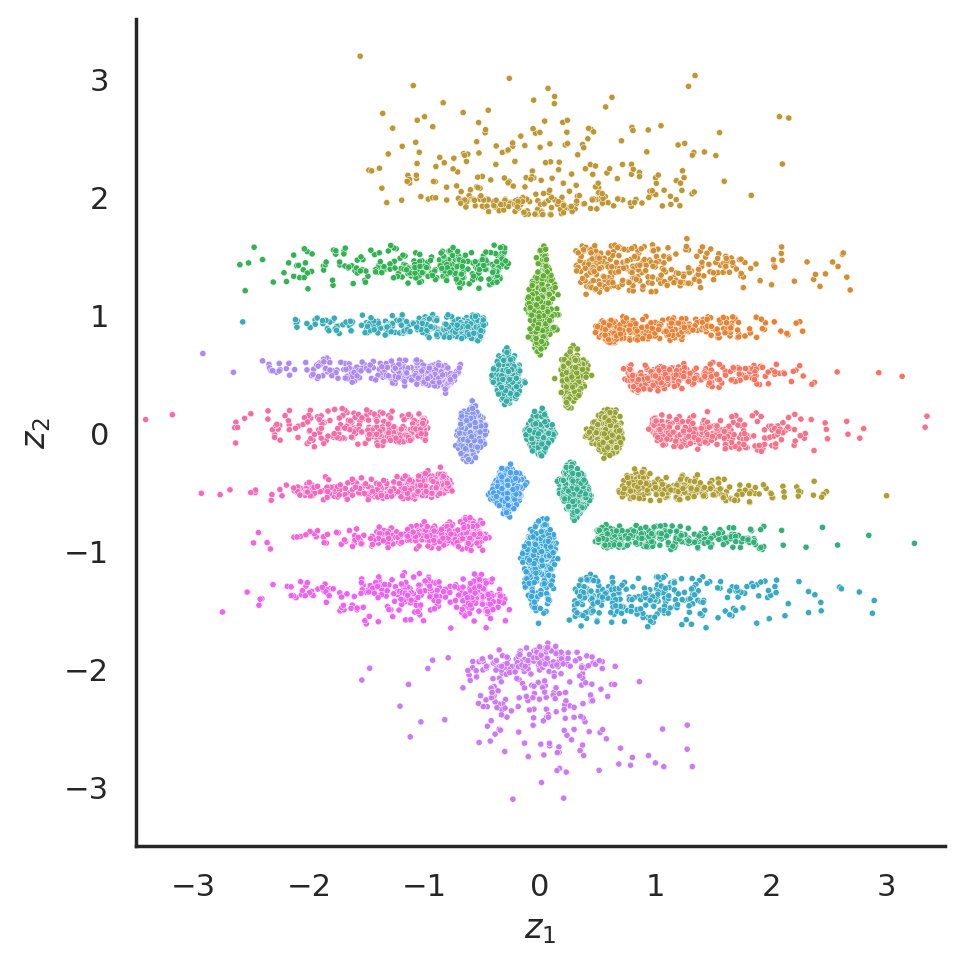}}
  \subfigure{\includegraphics[width=0.17\textwidth]{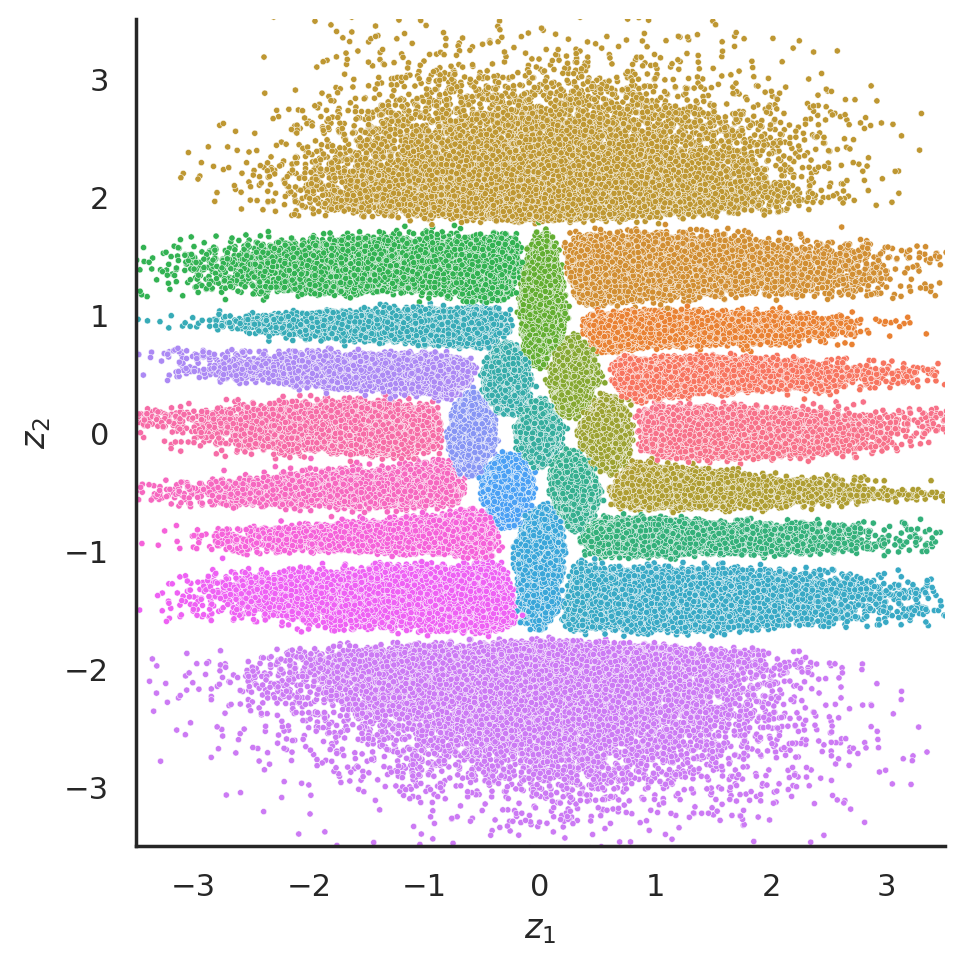}}
  \subfigure{\includegraphics[width=0.17\textwidth]{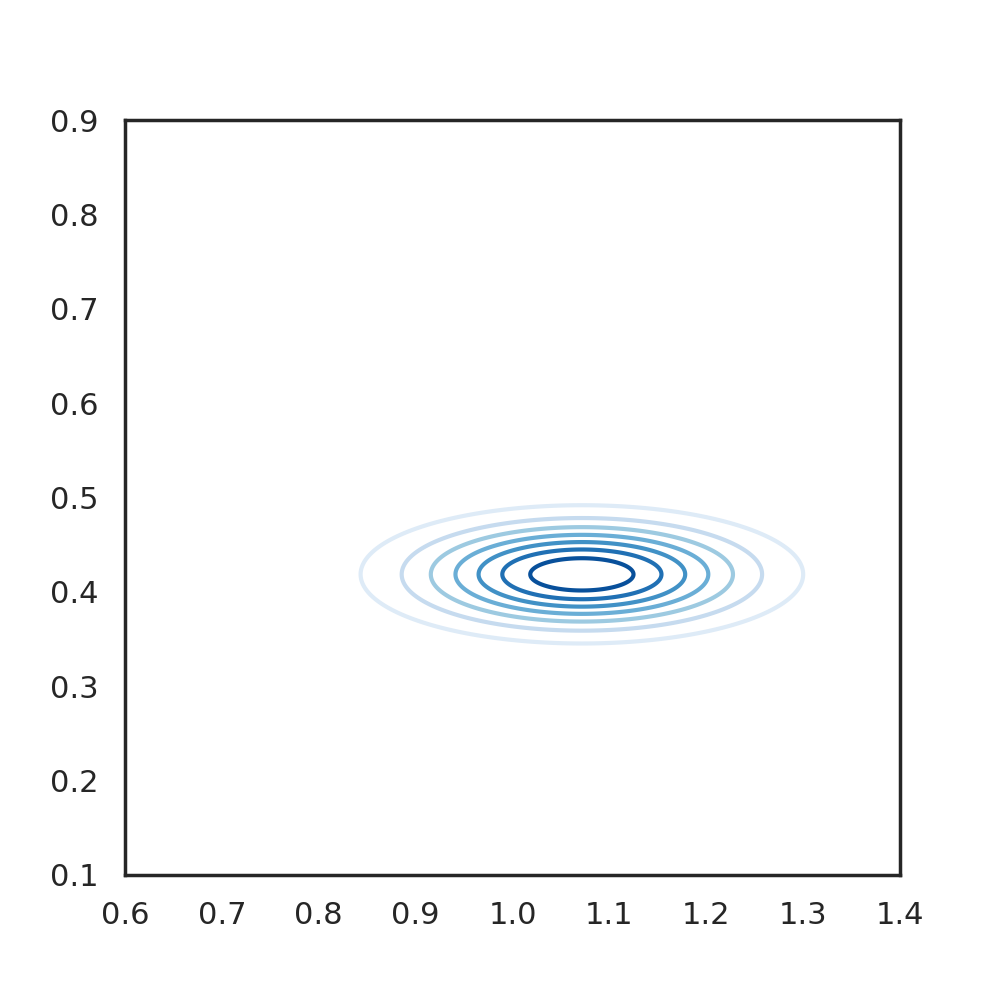}}

  \vspace{-5mm}
  \raisebox{0.5in}{\rotatebox[origin=t]{90}{C-VAE}} 
  \subfigure{\includegraphics[width=0.17\textwidth]{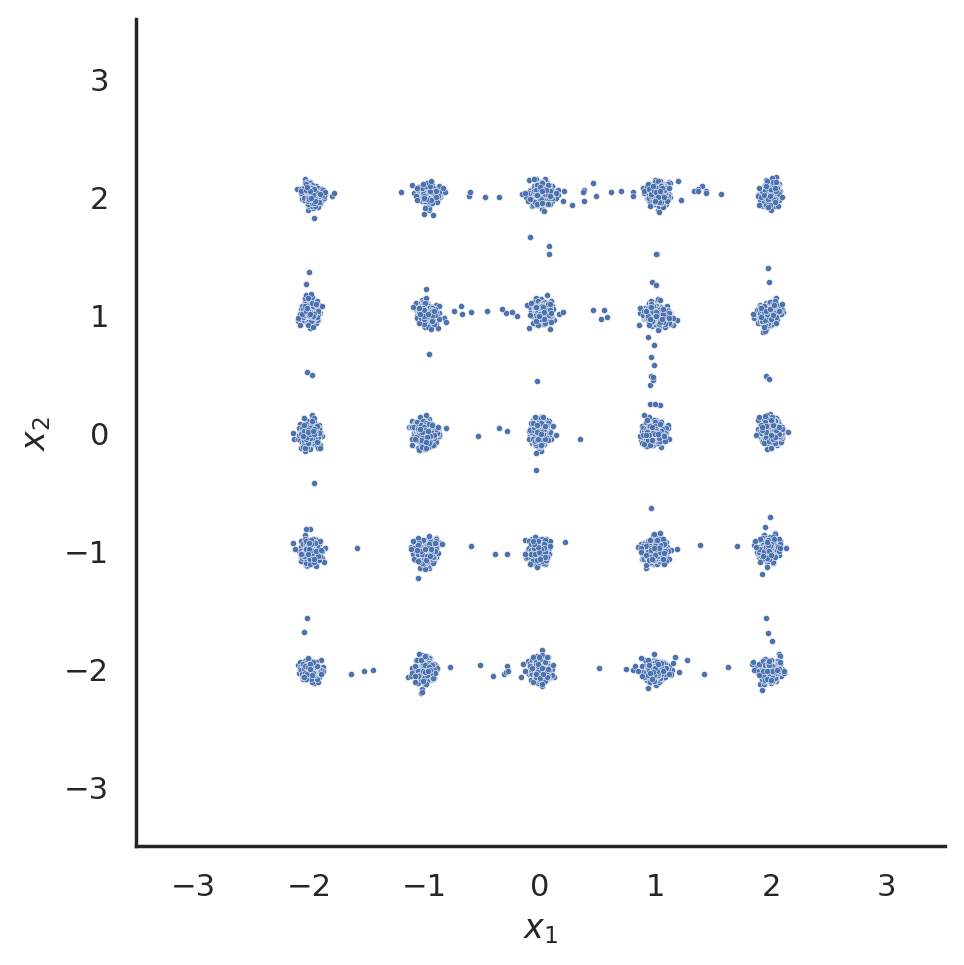}}
  \subfigure{\includegraphics[width=0.17\textwidth]{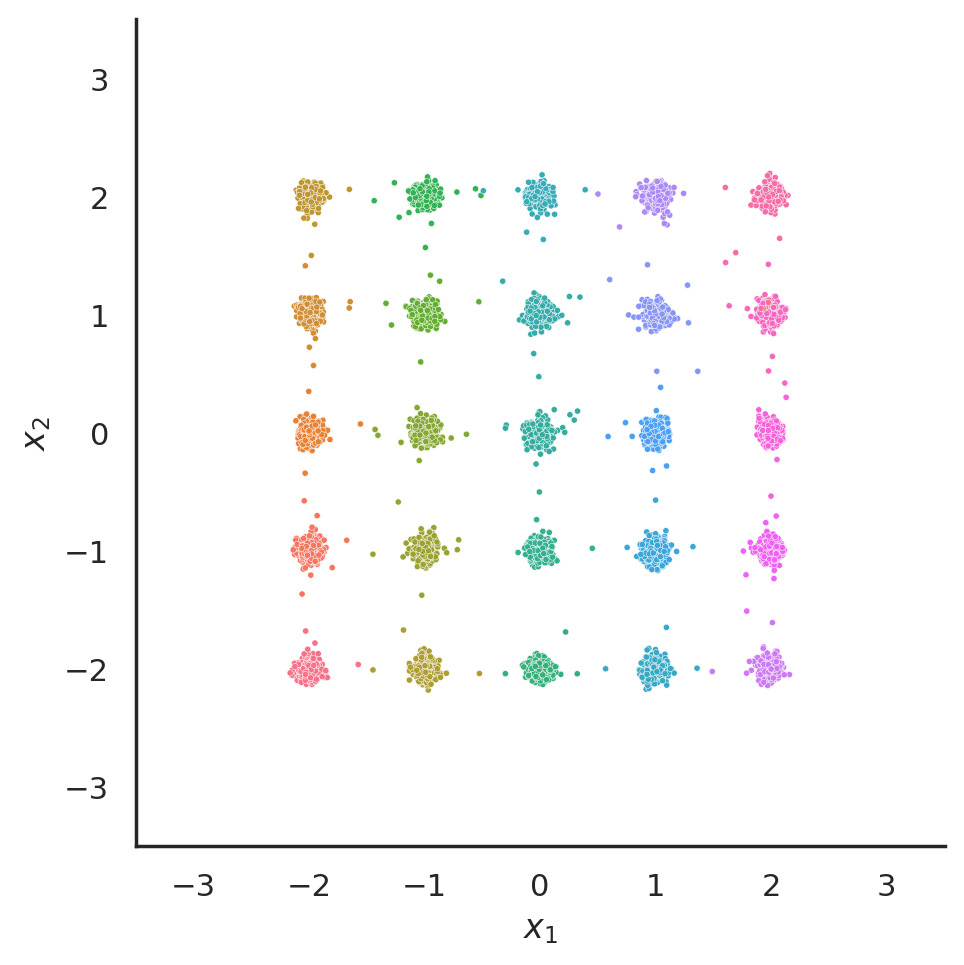}}
  \subfigure{\includegraphics[width=0.17\textwidth]{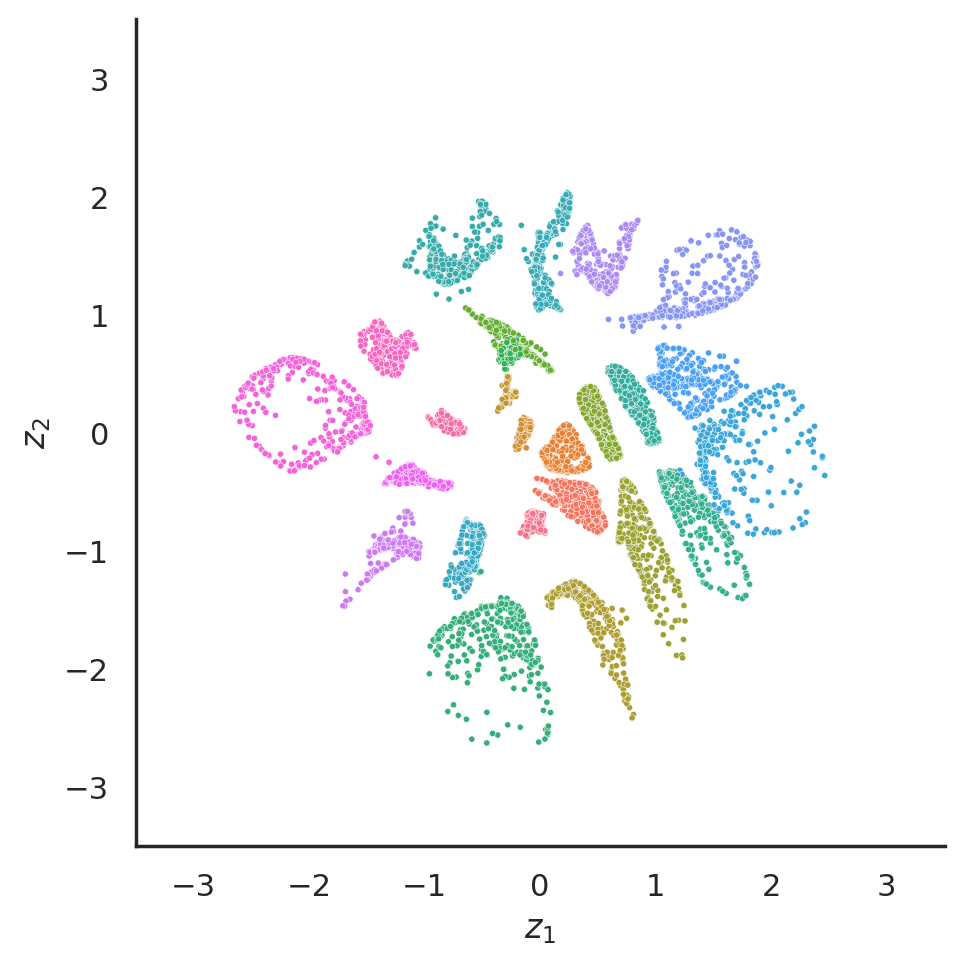}}
  \subfigure{\includegraphics[width=0.17\textwidth]{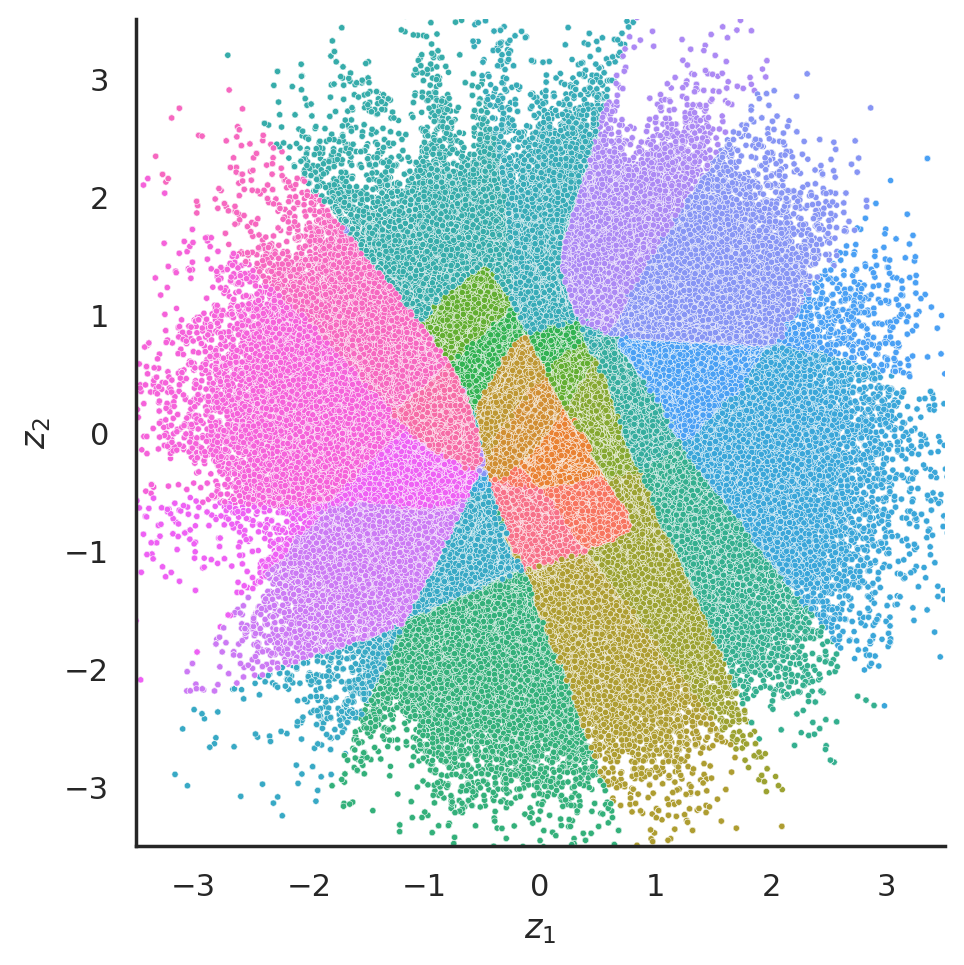}}
  \subfigure{\includegraphics[width=0.17\textwidth]{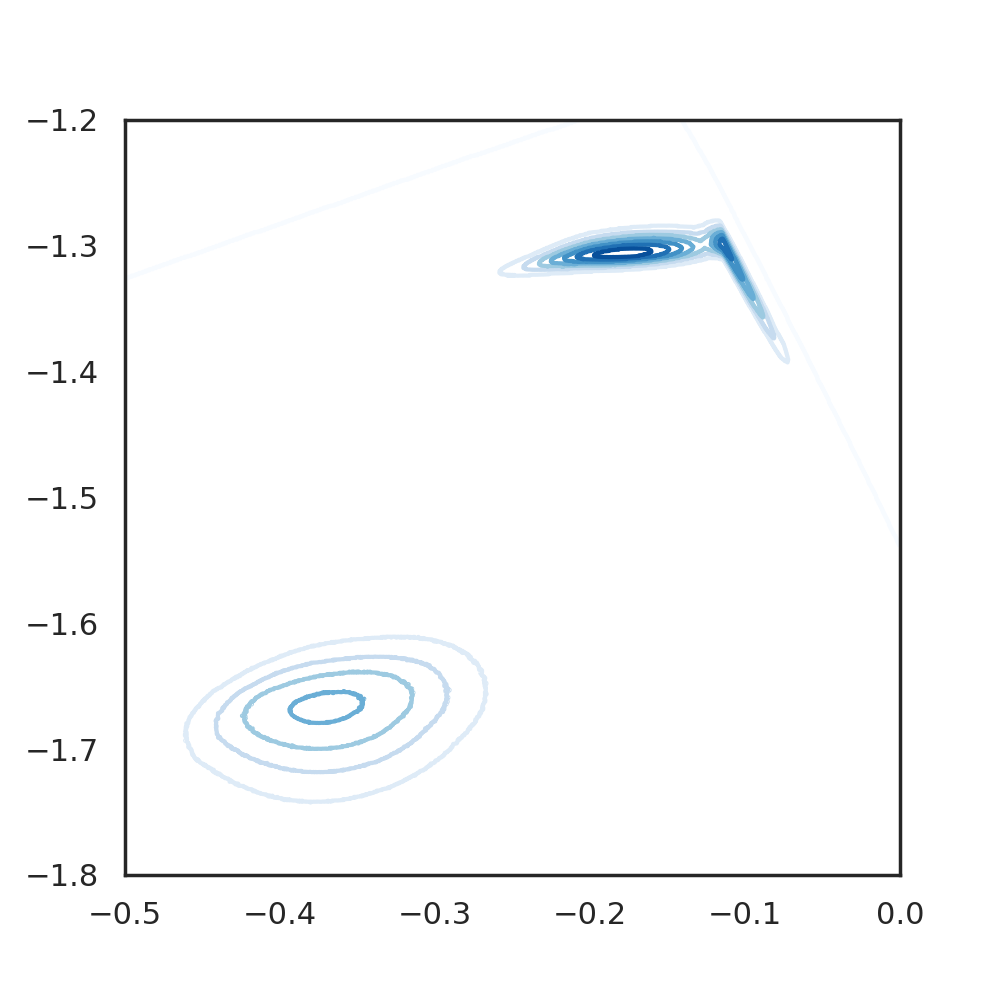}}
  
  \vspace{-5mm}
  \caption{Columns from left to right represent (1) random samples, (2) reconstructions, (3) latent representations, (4) samples from aggregated posteriors and (5) posterior, respectively. Each row represents a model. Colors correspond to the classes. We can see that C-VAE has the best sample quality (1), the prior hole problem is resolved in C-VAE (4),  C-VAE and VAE-NF can have non-isotropic-Gaussian posterior in contrast to Gaussians posteriors in others but VAE-NF is still approximately Gaussian-like (5).}
 
\label{fig:gaussians}
\end{figure*}

\begin{wrapfigure}{r}{0.19\textwidth}
  \begin{center}
    \includegraphics[width=0.18\textwidth]{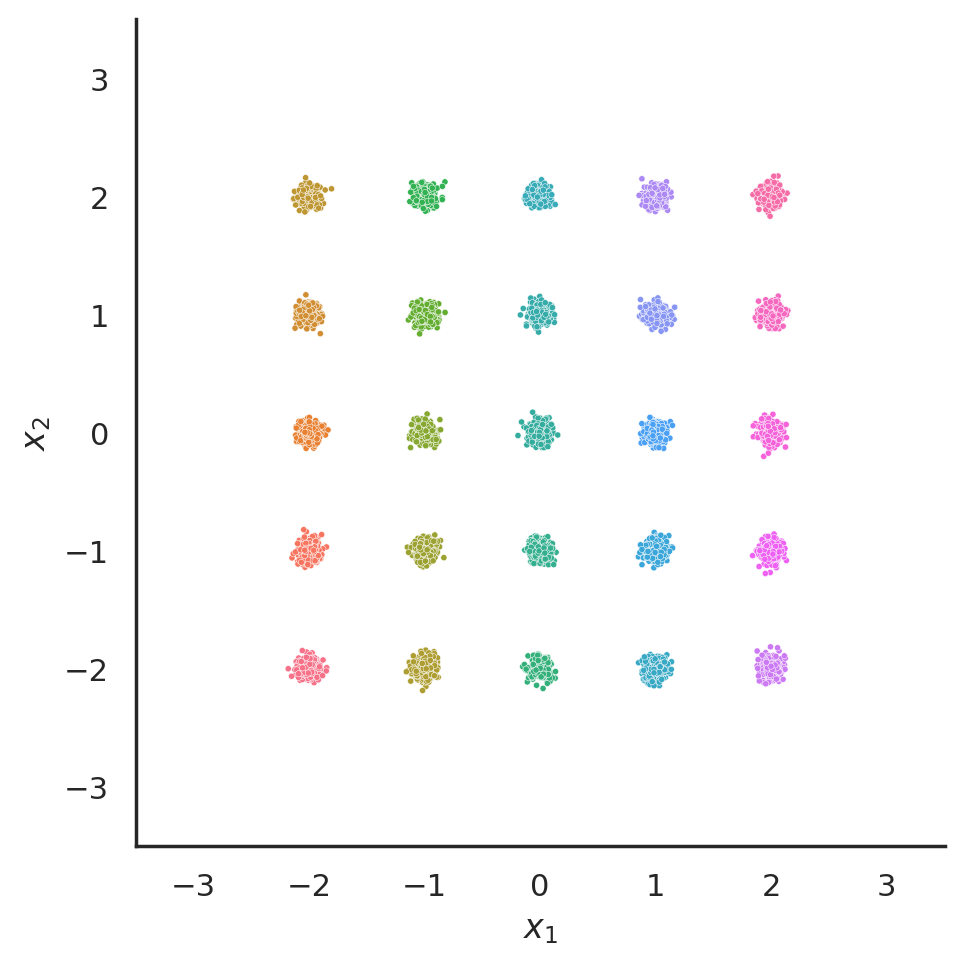}
  \end{center}
  \caption{Mixture of 25 Gaussians in $\mathbb{R}^2$.}
\end{wrapfigure}

\textbf{Mixture of Gaussians.} We first test our model on a two-dimensional synthetic dataset consisting of a mixture of 25 isotropic Gaussian distributions laid out on a grid \citep{Dumoulin2016AdversariallyLI}. The means are on the grids $\mu \in \{-2,1,0,1,2\}$ and we used a standard deviation $\sigma = 0.05$. Despite being a 2D toy example, it is not an easy task since the distribution defined there exhibits many modes that are separated by large low-density areas. In all experiments, we generated 300 samples from each Gaussian and synthesize a dataset with a total of 7500 data points. For all encoder and decoder neural networks, we used four fully connected layers with ReLU activations with Batch normalizations in between. All hidden layers have 256 neurons and we choose the latent dimension $d_z = 2$ to plot the latent space.  The prior is a 2D Gaussian with a mean of 0 and a standard deviation of 1, i.e., $p(z) = \gN(\mathbf{0},\mathbf{I_d})$. We pick the best model based on hyperparameter searching.

\textbf{Visualization.} To better understand how our model performs in generative tasks and latent representation learning, we visualize the random samples from models along with the training reconstructions. The samples are obtained by generating a latent code from $p(z)$ first and then decoding it by $p_\theta(x|z)$. We also show the latent representation learned from the whole dataset, the aggregate posterior of the model, and the posterior distribution for the individual data point. In order to visualize the encoding distribution, we choose the mean of the encoding distribution as latent representations. Aggregated posterior is produced by ancestral sampling of $q(z|x)p_D(x)$.
Figure~\ref{fig:gaussians} displays the results of all models in this experiment. We observe: 
\begin{enumerate}[noitemsep,topsep=0pt]
  \item For the data reconstruction (column 2), all models do well. But VAE, VAE-NF, WAE, and InfoVAE all generate samples (column 1) lying on the edges and inside between different components of mixture Gaussians, which implies the learned model distribution $P_\theta$ fails to match the data distribution $P_D$, which lives only on the grids.
  \item All models have learned well-separated representations (column 3). Prior holes---mismatch of aggregated posterior and prior---are visible for VAE, VAE-NF, WAE, and InfoVAE (column 4). Compared with VAE,  InfoVAE has smaller holes and better random sample quality. The difference between two models is the explicit penalty term on the discrepancy of $q(z)$ and $p(z)$, which is consistent with our hypothesis about the importance of closing the holes inside the prior. C-VAE outperforms InfoVAE by implicitly having an infinite penalty on the gap. WAE ends with an almost-deterministic encoder as the $ \sigma^2_\phi(x) \approx \mathbf{0}$ for all $x$. This makes the holes even larger than VAE.
  \item The last column of \figref{fig:gaussians} is the density plots of posteriors of a single data point. VAE, WAE, and InfoVAE all have Gaussian posteriors by assumption. WAE's posterior collapsed to a point. VAE-NF shows a non-isotropic-Gaussian posterior. It's clear that the posterior has non-zero covariance between $z_1$ and $z_2$. But overall it's still close to Gaussian with single mode. C-VAE presents a very different posterior which has multiple modes.
\end{enumerate}
\vspace{-3mm}
\begin{table}[ht]
\caption{Quantitative results on the mixture of Gaussians}
\vspace{0.1in}
\label{gaussian-table}
%\vskip 0.15in
\begin{center}
\begin{small}
\begin{sc}
\begin{tabular}{lcc}
\toprule
Model & High density ratio & std w/i modes\\
\midrule
VAE    & 71.5$\pm$ 0.005& 0.0778$\pm$ 0.0004 \\
VAE-NF   & 53.6$\pm$ 0.006& 0.0833$\pm$ 0.0006 \\
WAE & 60.8$\pm$ 0.006& 0.0764$\pm$ 0.0005 \\
InfoVAE    & 86.0$\pm$ 0.004& 0.0673$\pm$ 0.0004\\
\midrule
C-VAE   &\textbf{98.5$\pm$ 0.001} & \textbf{0.0478$\pm$ 0.0003} \\
\bottomrule
\end{tabular}
\end{sc}
\end{small}
\end{center}
\vskip -0.1in
\label{Tab:Q}
\end{table}

\vspace{-3mm}

\begin{table}[ht]
\caption{MMD between prior and aggregate posterior}
\vspace{0.1in}
\label{mmd-table}
%\vskip 0.15in
\begin{center}
\begin{small}
\begin{sc}
\begin{tabular}{lcc}
\toprule
Model & MMD\\
\midrule
VAE    & 0.0106$\pm$ 0.0004\\
VAE-NF    & 0.0161$\pm$ 0.0005\\
WAE & 0.0060$\pm$ 0.0002 \\
InfoVAE    & 0.0045$\pm$ 0.0002\\
\midrule
C-VAE   &\textbf{0.0032$\pm$ 0.0005} \\
\bottomrule
\end{tabular}
\end{sc}
\end{small}
\end{center}
\label{Tab:mmd}
\end{table}

\vspace{-2mm}

\begin{figure}[!hb]
  \centering
  \subfigure[VAE]{\includegraphics[width=0.15\textwidth]{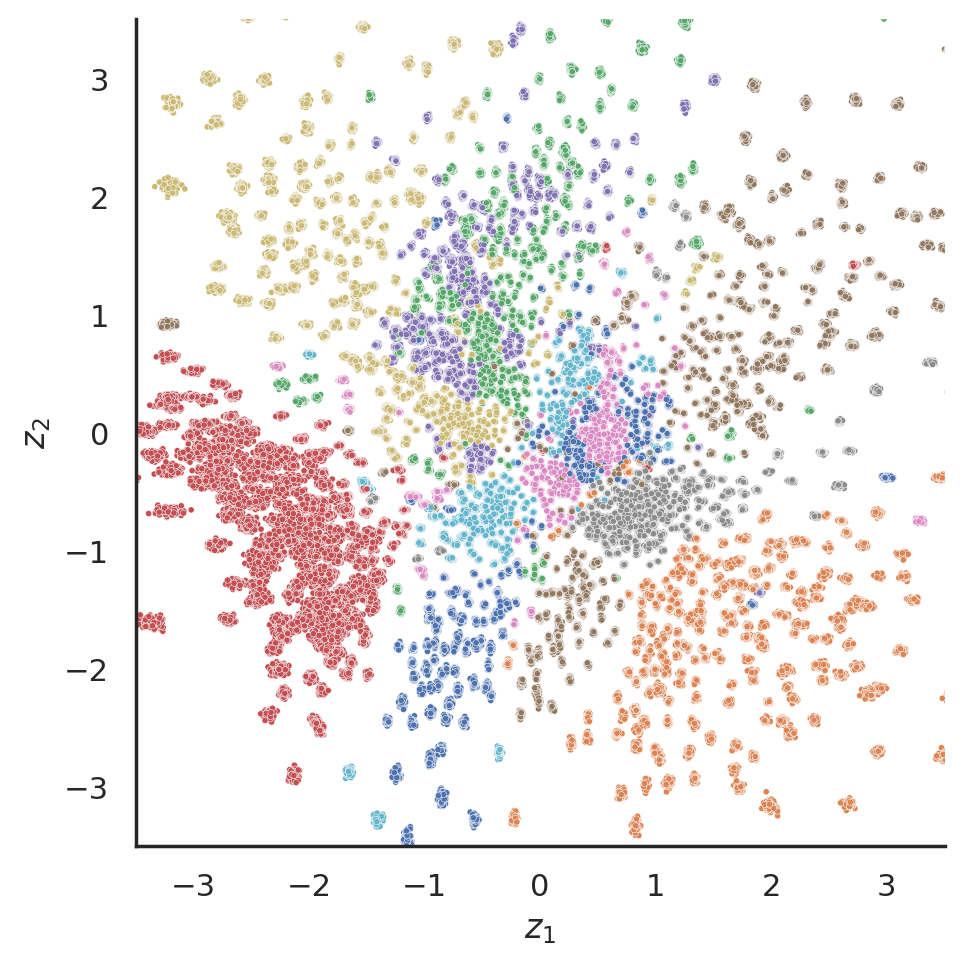}}
  \subfigure[VAE-NF]{\includegraphics[width=0.15\textwidth]{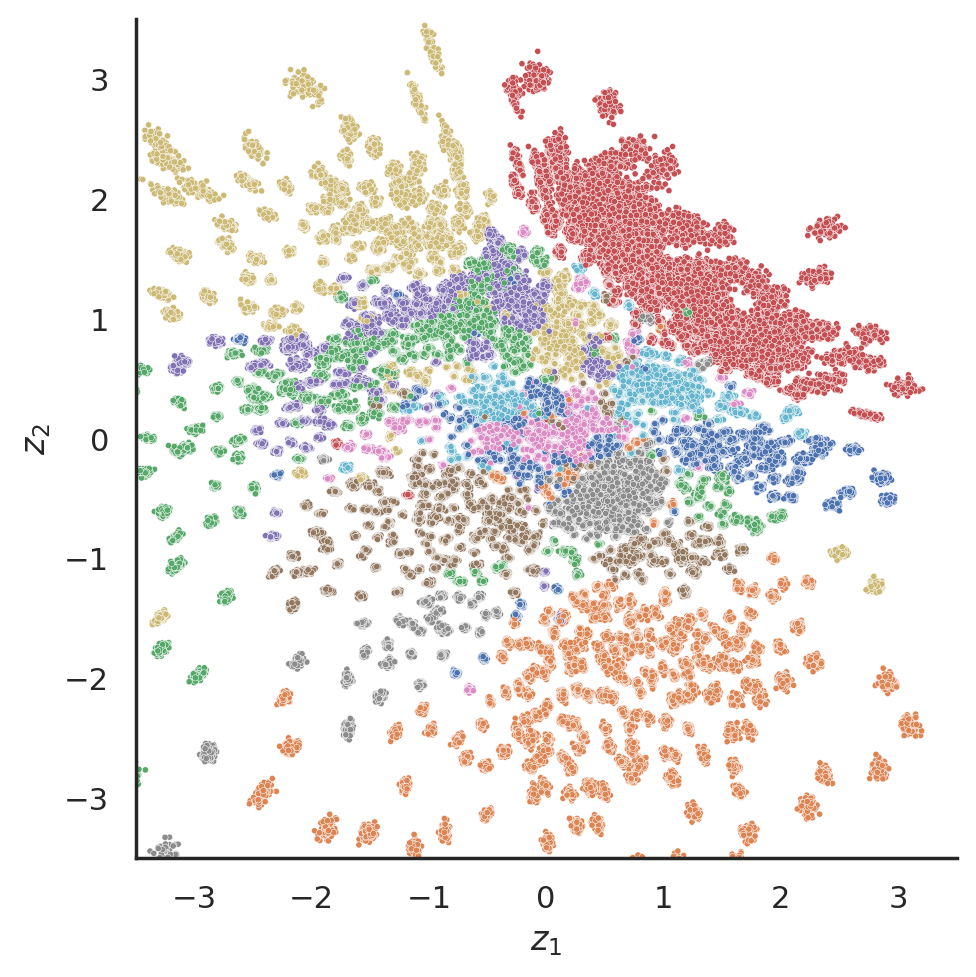}}
  
  \subfigure[WAE]{\includegraphics[width=0.15\textwidth]{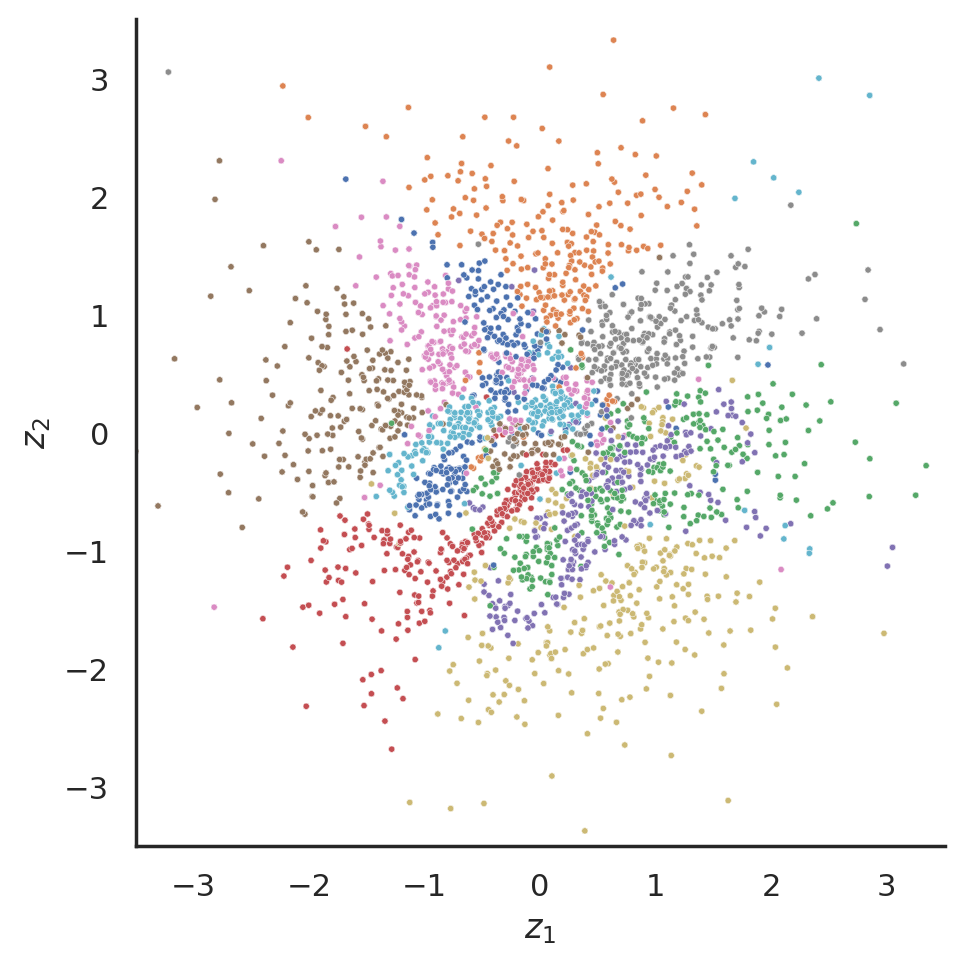}}
  \subfigure[InfoVAE]{\includegraphics[width=0.15\textwidth]{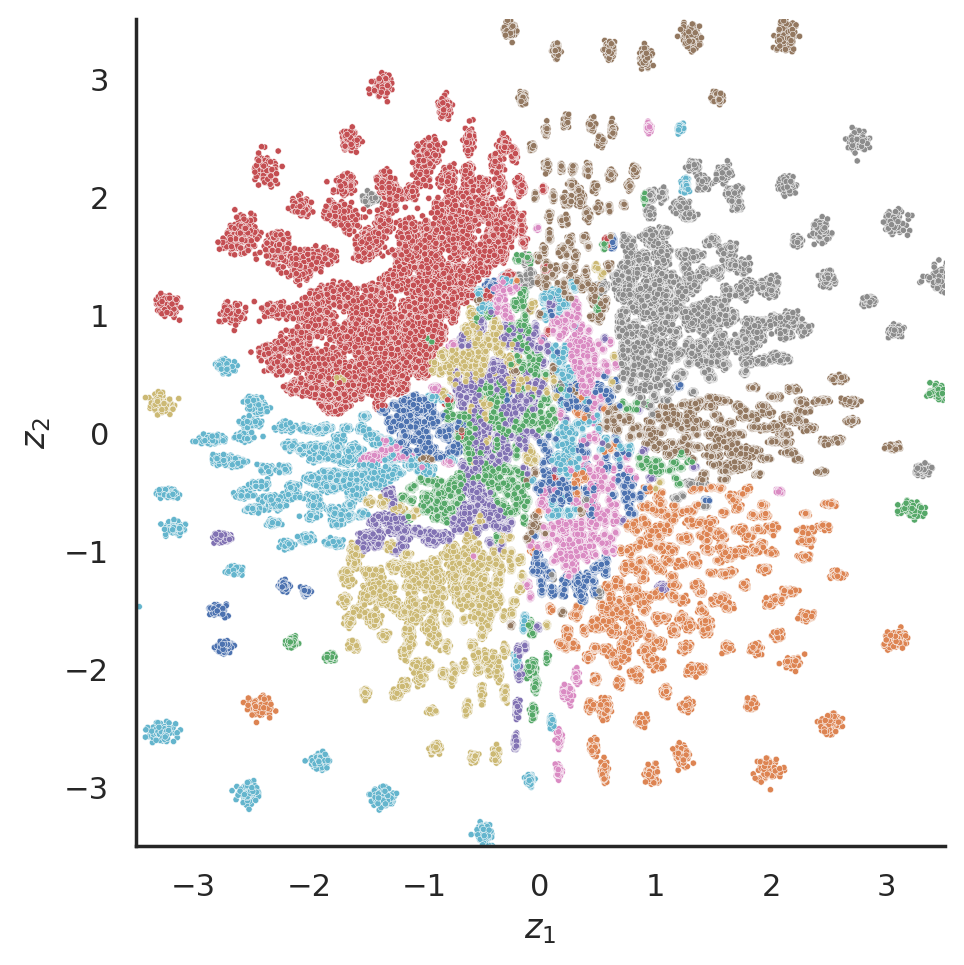}}
  \subfigure[C-VAE]{\includegraphics[width=0.15\textwidth]{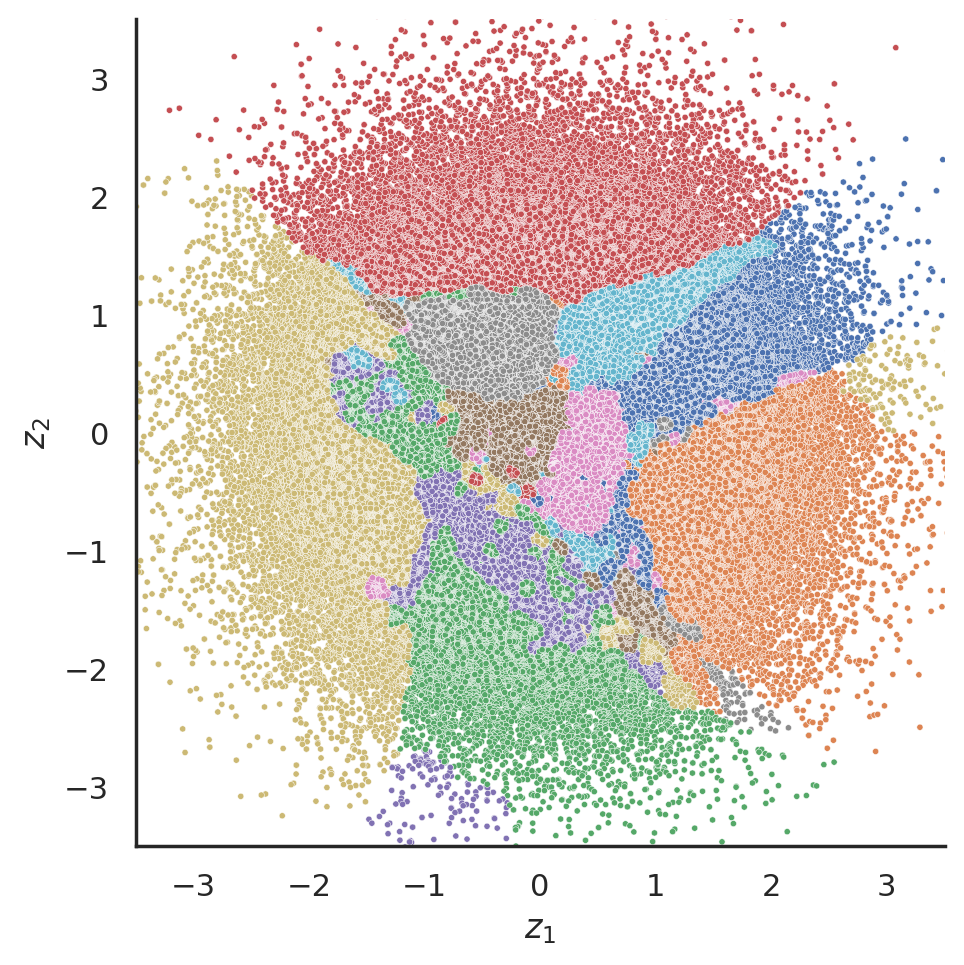}}
\vspace{-3mm}
  \caption{Samples from aggregated posterior $q(z)$. Each color represents a digit. C-VAE matches $p(z)$ best among the 5 models.}
 
\label{fig:mnist_qsample}
\end{figure}

\textbf{Quantitative results.} To measure performance quantitatively, we follow the metrics used in \citep{Azadi2018DiscriminatorRS}. We will assign each sample to its closest mixture component. A sample is considered as a high-density sample if it is within four standard deviations of its closest mixture component. The fraction of high-density samples in all samples will be calculated. Table \ref{Tab:Q} shows that C-VAE has attained the best high density ratio and the standard deviation of the learned model is 0.0478 which is closest to the ground truth of 0.05. We reported the Maximum Mean Discrepancy (MMD) between prior and aggregate posterior for all models in Table \ref{Tab:mmd}. C-VAE has the smallest MMD.

\begin{figure}[!hb]
  \centering
  % \rotatebox[origin=c]{90}{\bfseries Model 1\strut}
  
  \raisebox{0.5in}{\rotatebox[origin=t]{90}{VAE}} 
  \subfigure{\includegraphics[width=0.15\textwidth]{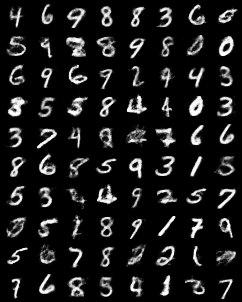}}
  \subfigure{\includegraphics[width=0.15\textwidth]{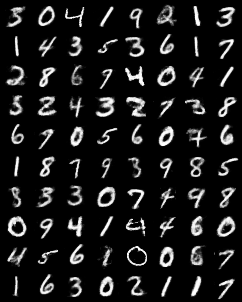}}
  \subfigure{\includegraphics[width=0.15\textwidth]{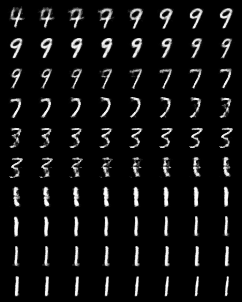}}
  
  \vspace{-3mm}
  \raisebox{0.5in}{\rotatebox[origin=t]{90}{VAE-NF}} 
  \subfigure{\includegraphics[width=0.15\textwidth]{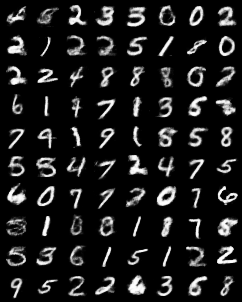}}
  \subfigure{\includegraphics[width=0.15\textwidth]{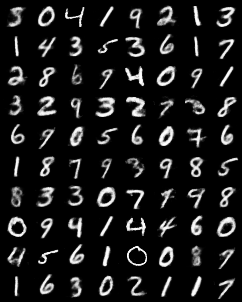}}
  \subfigure{\includegraphics[width=0.15\textwidth]{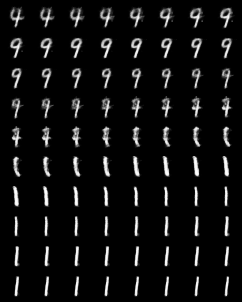}}
  
  \vspace{-3mm}
  \raisebox{0.5in}{\rotatebox[origin=t]{90}{WAE}} 
  \subfigure{\includegraphics[width=0.15\textwidth]{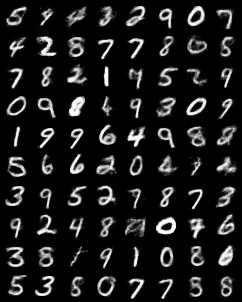}}
  \subfigure{\includegraphics[width=0.15\textwidth]{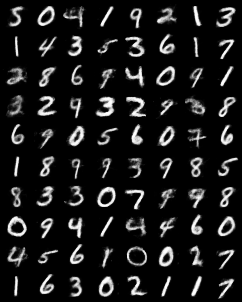}}
  \subfigure{\includegraphics[width=0.15\textwidth]{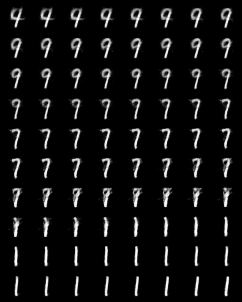}}

  \vspace{-3mm}
  \raisebox{0.5in}{\rotatebox[origin=t]{90}{InfoVAE}} 
  \subfigure{\includegraphics[width=0.15\textwidth]{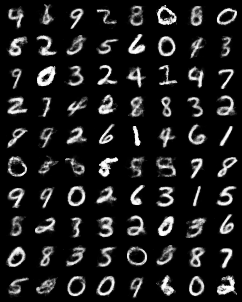}}
  \subfigure{\includegraphics[width=0.15\textwidth]{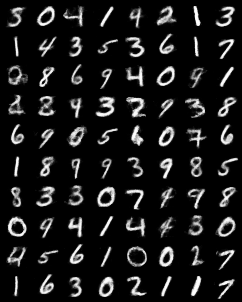}}
  \subfigure{\includegraphics[width=0.15\textwidth]{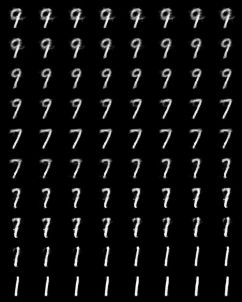}}

  \vspace{-3mm}
  \raisebox{0.5in}{\rotatebox[origin=t]{90}{C-VAE}} 
  \subfigure{\includegraphics[width=0.15\textwidth]{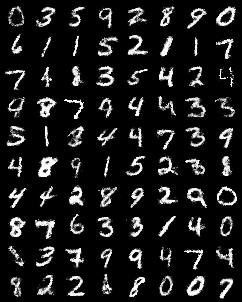}}
  \subfigure{\includegraphics[width=0.15\textwidth]{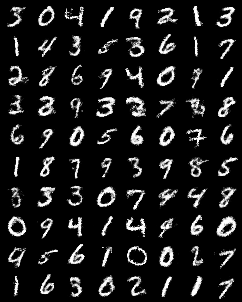}}
  \subfigure{\includegraphics[width=0.15\textwidth]{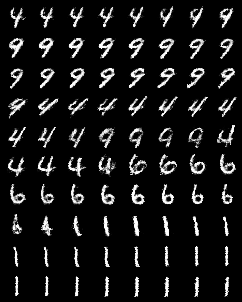}}

  \caption{columns from left to right represent (1) random samples, (2) reconstructions, (3) interpolations. Each row represents a model. The quality of random samples generated by C-VAE is better in the sense that samples are sharper and there are fewer samples that look like an average of digits. In the right column, we could see the decoded images of  linear interpolation in the latent space. The C-VAE interpolation is smoother and more realistic.}
 
\label{fig:mnist}
\end{figure}

\textbf{MNIST.} We then run our model on a subset of MNIST which consists of 2560 different hand-written digits images from 10 classes. We again compare our model with VAE, VAE-NF, WAE, and InfoVAE on similar tasks of synthetic examples. We also examine the interpolation in the latent space of all models. Latent dimension $d_z = 2$. All the autoencoder networks have the same architecture. We use convolutional layers paired with ReLU as the building block for the encoding/decoding networks. We choose Bernoulli likelihood in the experiments. Models are trained through Adam optimizer with $\beta_1 = 0.9$ and $ \beta_2 = 0.999$.

The results are reported in \figref{fig:mnist_qsample} and \figref{fig:mnist}. As expected, the quality of random samples and interpolation in latent space depends on how accurately $q(z)$ matches $p(z)$. \figref{fig:mnist_qsample} shows that C-VAE eliminates holes in the prior without hurting the performance of the model. WAE cannot match $p(z)$ perfectly because the encoder is collapsed to a deterministic map, which means $q(z)$ will have support on a finite number of $z$. The number is given by the number of data in the training set. InfoVAE mitigates but does not resolve the hole problem. Matching $q(z)$ to $p(z)$ better will need a larger penalty on the discrepancy, which will trade off against generative capacity. Interpolations, reconstructions, and samples are in \figref{fig:mnist}. 

% \begin{figure}
%   \centering
  
%   \subfigure[VAE]{\includegraphics[width=0.17\textwidth]{./figures/VAE.png}}
%   \subfigure[WAE]{\includegraphics[width=0.17\textwidth]{./figures/InfoVAE0_2.png}}
%   \subfigure[InfoVAE]{\includegraphics[width=0.17\textwidth]{./figures/InfoVAE1_2.png}}
%   \subfigure[C-VAE]{\includegraphics[width=0.17\textwidth]{./figures/CVAE_samples_mnist.png}}

%   \caption{Random samples generated from prior}

% \end{figure}

\section{Conclusion}

By recasting VAE as EOT, we introduced an EOT-based training scheme for latent variable models, which enables flexible posterior approximation and prior selection. Our model resolved the prior hole problem naturally by EOT. We verify our claims on synthetic mixture of Gaussians dataset and MNIST.

\section*{Acknowledgements}
This work was supported in part by DARPA Sail-on W911NF2020001, ASIST W912CG22C0001, and NSF MRI 2117429. The views and conclusions contained in this document are those of the authors and should not be interpreted as representing the official policies, either expressed or implied, of the DARPA or ARO, or the U.S. Government.


\bibliography{CVAE}
\bibliographystyle{icml2023}

%%%%%%%%%%%%%%%%%%%%%%%%%%%%%%%%%%%%%%%%%%%%%%%%%%%%%%%%%%%%%%%%%%%%%%%%%%%%%%%
%%%%%%%%%%%%%%%%%%%%%%%%%%%%%%%%%%%%%%%%%%%%%%%%%%%%%%%%%%%%%%%%%%%%%%%%%%%%%%%
% APPENDIX
%%%%%%%%%%%%%%%%%%%%%%%%%%%%%%%%%%%%%%%%%%%%%%%%%%%%%%%%%%%%%%%%%%%%%%%%%%%%%%%
%%%%%%%%%%%%%%%%%%%%%%%%%%%%%%%%%%%%%%%%%%%%%%%%%%%%%%%%%%%%%%%%%%%%%%%%%%%%%%%
% \newpage
% \appendix
% \onecolumn

% You can have as much text here as you want. The main body must be at most $8$ pages long.
% For the final version, one more page can be added.
% If you want, you can use an appendix like this one, even using the one-column format.
%%%%%%%%%%%%%%%%%%%%%%%%%%%%%%%%%%%%%%%%%%%%%%%%%%%%%%%%%%%%%%%%%%%%%%%%%%%%%%%
%%%%%%%%%%%%%%%%%%%%%%%%%%%%%%%%%%%%%%%%%%%%%%%%%%%%%%%%%%%%%%%%%%%%%%%%%%%%%%%

\end{document}